\documentclass[sigconf]{acmart}

\usepackage{amsmath}
\usepackage{multirow}
\usepackage{makecell}
\usepackage{subcaption}
\usepackage{array}
\usepackage{xcolor}
\usepackage[x11names]{xcolor}  
\usepackage{colortbl}       
\usepackage{balance}
\AtBeginDocument{%
  }

\copyrightyear{2026}
\acmYear{2026}
\setcopyright{cc}
\setcctype{by}
\acmConference[MM '26] {Proceedings of the 34th ACM International Conference on Multimedia}{November 10--14, 2026}{Rio de Janeiro, Brazil.}
\acmBooktitle{Proceedings of the 34th ACM International Conference on Multimedia (MM '26), November 10--14, 2026, Rio de Janeiro, Brazil}
\acmISBN{979-8-4007-2213-4/2026/11}
\acmDOI{10.1145/XXXXXX.XXXXXX}

\acmSubmissionID{3563}

\begin{document}

\title{Exo2EgoPose: Leveraging Exocentric Demonstrations for\\Vision-Language guided Egocentric 3D Hand Pose Forecasting}

\author{Zhaofeng Shi}
\email{zfshi@std.uestc.edu.cn}
\orcid{0000-0001-6313-8670}
\affiliation{
  \institution{University of Electronic Science and Technology of China}
  \city{Chengdu}
  \state{Sichuan}
  \country{China}
  \postcode{611731}
}

\author{Heqian Qiu}
\email{hqqiu@uestc.edu.cn}
\orcid{0000-0002-0963-0311}
\authornote{Corresponding authors.}
\affiliation{%
  \institution{University of Electronic Science and Technology of China}
  \city{Chengdu}
  \state{Sichuan}
  \country{China}
  \postcode{611731}
}

\author{Lanxiao Wang}
\email{lanxiaowang@uestc.edu.cn}
\orcid{0000-0002-3745-0262}
\affiliation{%
  \institution{University of Electronic Science and Technology of China}
  \city{Chengdu}
  \state{Sichuan}
  \country{China}
  \postcode{611731}
}

\author{Xiang Li}
\email{xianglee@std.uestc.edu.cn}
\orcid{0009-0009-1196-2276}
\affiliation{%
 \institution{University of Electronic Science and Technology of China}
  \city{Chengdu}
  \state{Sichuan}
  \country{China}
  \postcode{611731}
}

\author{Hongliang Li}
\email{hlli@uestc.edu.cn}
\orcid{0000-0002-7481-095X}
\affiliation{%
  \institution{University of Electronic Science and Technology of China}
  \city{Chengdu}
  \state{Sichuan}
  \country{China}
  \postcode{611731}
}


\renewcommand{\shortauthors}{Zhaofeng Shi, Heqian Qiu, Lanxiao Wang, Xiang Li, and Hongliang Li}

\begin{abstract}
Perceiving multimodal cues and forecasting fine-grained actions from an egocentric (Ego) perspective is vital for applications like robot manipulation. However, previous studies either rely mainly on under-informed visual inputs to predict coarse human motions or follow the VRM/VLA paradigm, which suffers from insufficient robot data and the gap between human and robot embodiments. We observe that 3D hand pose naturally serves as a unified representation to bridge human-robot actions. Hence, we investigate an under-explored Vision-Language guided Egocentric 3D Hand Pose Forecasting (VL-EHPF) task, which aims to predict future Ego 3D hand poses from visual observations, a language instruction, and pose states. To overcome the limited field-of-view and highly dynamic motions in the Ego view, we propose a framework dubbed Exo2EgoPose, which innovatively leverages holistic and stable exocentric (Exo) demonstrations as guidance to compensate for partial and dynamic Ego-view cues. Specifically, we introduce a Dual-level Exocentric Reconstruction Module (DERM), which incorporates the paired Exo videos as supervision to reconstruct their video-level and chunked frame-level representations, thereby modeling spatial contexts and temporal dynamics. Then, the Global-to-Local Modulation Module (GLMM) utilizes the reconstructed hierarchical Exo representations for progressive feature refinement via attention mechanisms and adaptive modulation, enabling comprehensive Exo guidance for accurate Ego hand pose forecasting. Extensive experiments on \textit{AssemblyHands}, \textit{Ego-Exo4D}, and our newly constructed \textit{EgoMe-pose} benchmarks show the superiority of our method, which outperforms state-of-the-art methods by a large margin. Moreover, it demonstrates an effective human-to-robot transfer capability and yields improvements on the \textit{CALVIN} dataset. Code is available at \href{https://github.com/ZhaofengSHI/Exo2EgoPose}{https://github.com/ZhaofengSHI/Exo2EgoPose}.

\end{abstract}



\maketitle

\section{Introduction}

Perceiving the current state within the embodied space and forecasting fine-grained future actions from an egocentric (Ego) perspective plays a pivotal role in multimedia AI systems, which can be extended to diverse applications such as augmented reality (AR) \cite{shi2023dual,pei2022hand,qian2022arnnotate} and intelligent robot manipulation \cite{duan2022survey,zheng2026egoscale,kadalagere2023review,ji2025robobrain}. Such a remarkable capability not only provides profound insights into the cognitive processes underlying human intentions \cite{kosch2023survey,mcclelland2022capturing,shi2024cross}, but also possesses the potential to transfer knowledge from high-level human activities to downstream robotic execution tasks \cite{dong2023novel,li2023human,lee2024human}.

Hands are the primary medium through which humans interact with the physical world. This has driven extensive interest in the Ego hand pose, which provides an explicit representation of human activities and naturally serves as a bridge between human and robot actions. Many efforts focus on estimating the 2D/3D hand poses \cite{ohkawa2023assemblyhands,liu2024single,prakash20243d,moon2020interhand2,simon2017hand,grauman2024ego,pavlakos2024reconstructing} from the given Ego images or videos to interpret the current action state. To infer the next-step actions crucial for decoding human intentions, researchers make a step forward to predict future hand motions \cite{bao2023uncertainty,ma2025madiff,liu2022joint}. For example, the pioneering OCT \cite{liu2022joint} jointly predicts 2D hand motions and interactive hotspots. USST \cite{bao2023uncertainty} forecasts the Ego 3D hand trajectories via an uncertain state space model. MADiff \cite{ma2025madiff} makes improvements by adopting Diffusion-based \cite{ho2020denoising} strategies. However, these methods predict coarse-level trajectories or hotspots by relying mainly on the visual modality, which is under-informed without explicit and complete task contexts such as detailed pose states and textual instructions, thereby struggling to forecast fine-level joint-wise dynamics. Recently, numerous studies \cite{lynch2020learning,lynch2020language,mees2022matters,brohan2022rt,zitkovich2023rt,kim2024openvla} consider forecasting end-effector motions for robots following the trending Visual Robot Manipulation (VRM) or Vision-Language-Action (VLA) paradigms, which take diverse modality signals as input and output specific robot task executions. Despite the impressive achievements, they are hindered by the limited scale of robot data \cite{yuan2025embodied,niu2025pre}. Although some works \cite{wuunleashing,yang2025egovla,yoshida2025developing} attempt to incorporate human data to alleviate this problem, they still yield sub-optimal results due to the significant gap between human and robot embodiments \cite{li2025developments,zhou2025mitigating}.

\begin{figure}[!t]
\centering
\includegraphics[width=0.91\linewidth]{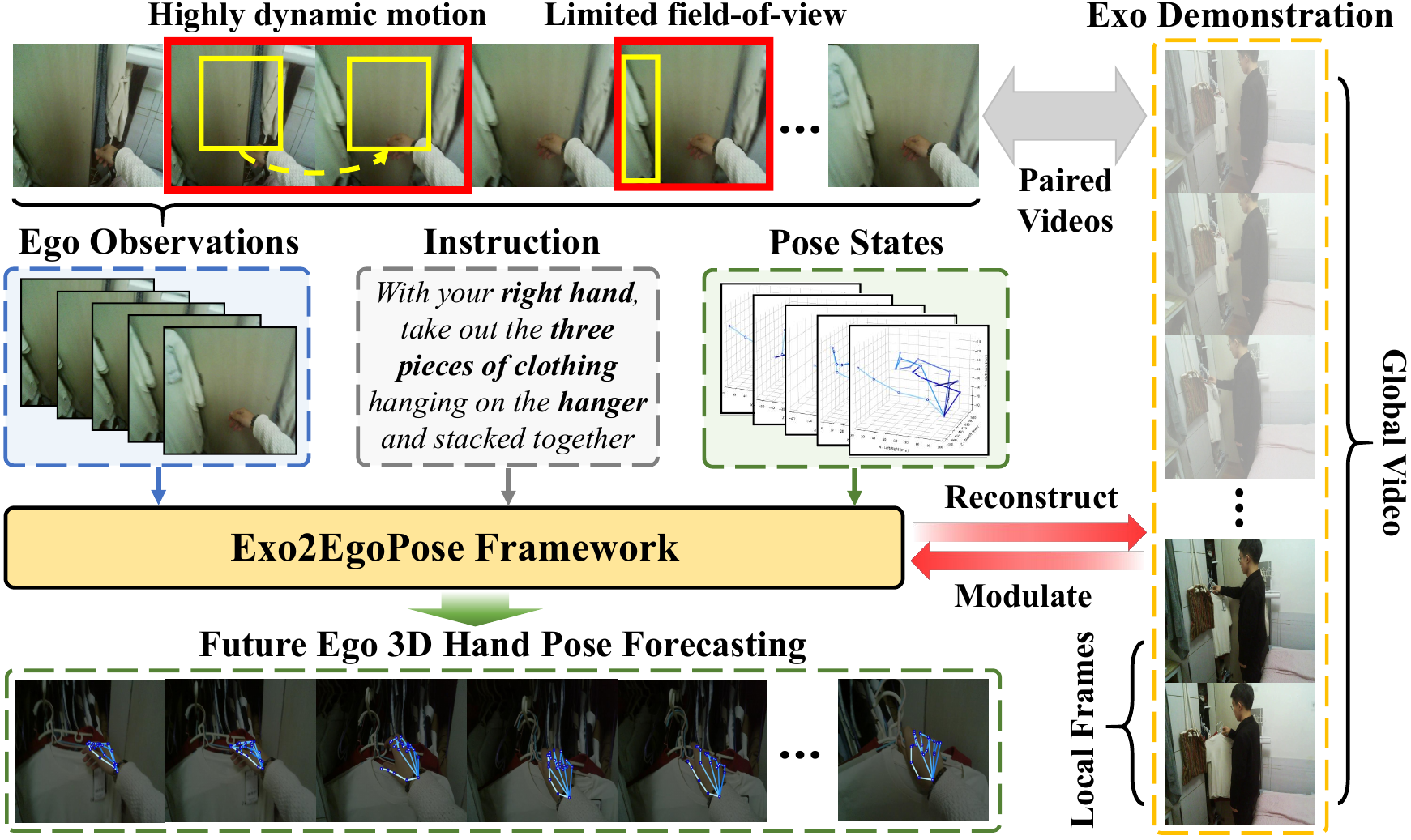}
\caption{Schematic of the VL-EHPF task and ideology of the proposed Exo2EgoPose Framework.}
\label{fig:1}
\vspace{-2mm}
\end{figure}

To this end, we investigate an under-explored Vision-Language guided Egocentric 3D Hand Pose Forecasting (VL-EHPF) task, which aims to forecast future Ego 3D hand poses based on input visual observations, a language instruction, and pose states. Although recent AR-VRM \cite{yang2025ar} and concurrent SFHand \cite{liu2025sfhand} propose pretraining-based and streaming-based approaches for the relevant tasks, they overlook the unique challenges in the Ego view (i.e., limited field-of-view and highly dynamic motions as shown in Fig. \ref{fig:1}). Specifically, on the one hand, Ego videos are recorded by head-mounted devices and only cover limited regions in front of the subject, resulting in insufficient contextual cues like truncated hand or object appearances, which remarkably affect the understanding of hand-object interaction and prediction of hand keypoints. On the other hand, freely moving Ego cameras coupled with rapid hand maneuvers lead to highly dynamic backgrounds and complex hand-background relative motions. Such dynamics make it difficult to accurately capture both current and upcoming hand motion patterns.

Inspired by previous adaptation \cite{liu2024single,ohkawa2025exo2egodvc,shi2025unsupervised,quattrocchi2024synchronization} or transfer \cite{huang2024egoexolearn,zhang2025exo2ego,shi2026test,shi2024cognition} works that utilize complementary information across exocentric (Exo) and egocentric (Ego) views, we propose a novel framework dubbed Exo2EgoPose to address the aforementioned problems, as shown in Fig. \ref{fig:1}. It is the first exploration to leverage holistic and stable Exo demonstration videos, which contain absolute locations and motion dynamics, serving as essential guidance to compensate for partial and dynamic Ego cues in 3D hand pose forecasting. In detail, to construct the Ego-Exo cross-view correspondence, we introduce a Dual-level Exocentric Reconstruction Module (DERM). It incorporates the paired Exo demonstration videos as supervision during training, aiming to reconstruct the video-level and chunked frame-level Exo representations extracted by an MAE \cite{he2022masked} encoder based on multimodal Ego inputs, which facilitates modeling complex spatial contexts and temporal dynamics in the Ego view. Then, we inject the reconstructed hierarchical Exo representations into the learned Ego features via the Global-to-Local Modulation Module (GLMM). This module performs progressive feature alignment and distribution calibration via attention mechanisms and adaptive modulation units (AMU) \cite{peebles2023scalable}, thereby fully exploiting the Exo-view guidance for accurate Ego 3D hand pose forecasting. For a comprehensive evaluation, we evaluate our Exo2EgoPose not only on the existing processed \textit{AssemblyHands} and \textit{Ego-Exo4D} benchmarks, but also on our newly constructed \textit{EgoMe-pose} benchmark based on the EgoMe dataset \cite{qiu2025egome}, which contains paired videos from the perspectives of observer and follower in real-life scenarios. Moreover, we also evaluate the human-to-robot knowledge transfer capability on the challenging robotic \textit{CALVIN} dataset, where our method achieves impressive performance improvements.

The major contributions can be concluded as follows:

\begin{itemize}

\item We investigate an under-explored Vision-Language guided Egocentric 3D Hand Pose Forecasting (VL-EHPF) task and propose a novel Exo2EgoPose framework, which leverages holistic and stable Exo demonstrations to compensate for partial and dynamic Ego cues in 3D hand pose forecasting.

\item We develop a Dual-level Exocentric Reconstruction Module (DERM) to model spatial contexts and temporal dynamics by reconstructing Exo representations at different levels during training. Moreover, a Global-to-Local Modulation Module (GLMM) uses the hierarchical Exo representations to perform progressive feature refinement via attention and AMU for accurate Ego 3D hand pose forecasting.

\item We construct a novel \textit{EgoMe-pose} benchmark. Experiments not only show that our Exo2EgoPose outperforms state-of-the-art methods by a large margin on \textit{AssemblyHands}, \textit{Ego-Exo4D}, and \textit{EgoMe-pose} benchmarks, but also indicate its human-to-robot transfer capability on the \textit{CALVIN} dataset.

\end{itemize}

\section{Related Work}

\begin{figure*}[!t]
\centering
\includegraphics[width=0.81\linewidth]{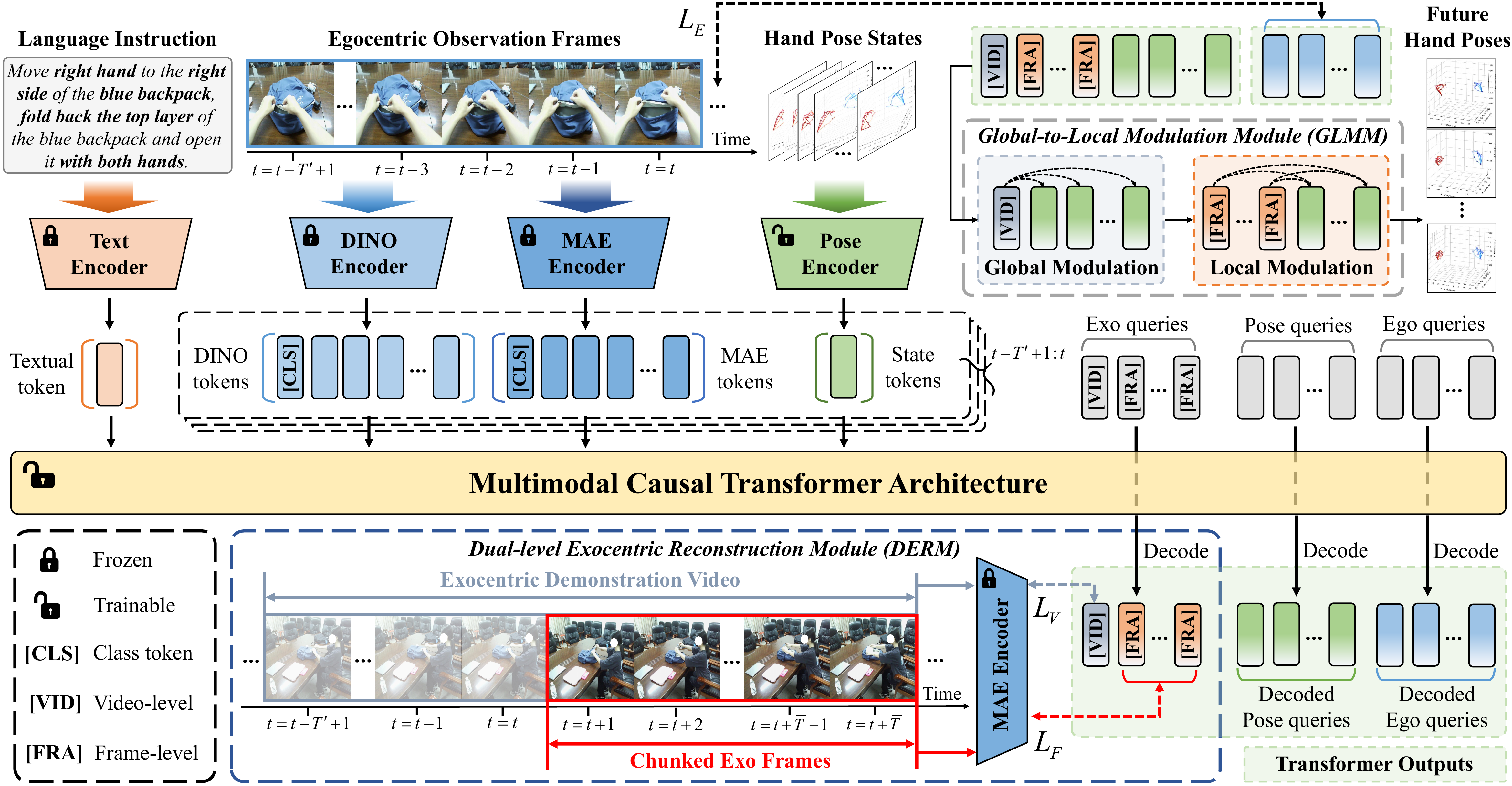}
\caption{Overview of our Exo2EgoPose framework. First, we adopt multiple modality-specific encoders to extract features, which are then tokenized and fed into a multimodal Transformer with initialized queries. Then, the DERM reconstructs the representations of the whole video clip and future chunked frames from the Exo perspective during training to construct Ego-Exo cross-view correspondence. Next, the GLMM utilizes the reconstructed Exo knowledge as guidance and performs comprehensive global-to-local feature refinement, which facilitates accurate prediction for the future Ego 3D hand poses.}
\label{fig:2}
\end{figure*}

\subsection{Ego-Exo Cross-view Understanding}

Following the introduction of Ego datasets \cite{sigurdsson2018charades,grauman2022ego4d,damen2022rescaling,de2009guide,banerjee2025hot3d,qi2025d3net}, researchers have increasingly focused on incorporating Exo videos to overcome the inherent challenges in the Ego view. In recent years, many Ego-Exo datasets \cite{grauman2024ego,li2024egoexo,qiu2025egome,huang2024egoexolearn,sener2022assembly101,ohkawa2023assemblyhands,kwon2021h2o,jia2020lemma} have been proposed. H2O \cite{kwon2021h2o} is a multi-view dataset for analyzing handovers. EgoExoLearn \cite{huang2024egoexolearn} aims to bridge asynchronous procedural activities at a semantic level. Assembly101 \cite{sener2022assembly101} is a multi-view dataset for assembling toys, while AssemblyHands \cite{ohkawa2023assemblyhands} further improves its pose annotations. Ego-Exo4D \cite{grauman2024ego} is currently the largest dataset with multi-view videos and diverse annotations. EgoMe \cite{qiu2025egome} contains Ego and Exo videos recorded from the observer and follower perspectives. Building upon EgoMe, we perform automatic 3D hand pose labeling and construct a brand-new \textit{EgoMe-pose} benchmark.

Meanwhile, many works \cite{shi2024cognition,shi2025unsupervised,luo2025viewpoint,ardeshir2018exocentric,li2021ego,xue2023learning,qiu2026ego} have been proposed to model Ego-Exo correlations for various tasks. These tasks can be summarized into three levels: video-level tasks for learning globally unified representations, such as cross-view association \cite{luo2025viewpoint,xue2023learning,huang2025sound} and video captioning \cite{xu2024retrieval,zhang2024self}; segment-level tasks for understanding procedural activities like action planning \cite{huang2024egoexolearn,zhang2025exo2ego} and fine-level video understanding \cite{shi2025unsupervised,quattrocchi2024synchronization}; frame-level tasks for parsing pixel-level patterns, such as novel view synthesis \cite{liu2020exocentric,liu2024exocentric} and pose estimation \cite{grauman2024ego,liu2024single}. Unlike the above tasks, we investigate the under-explored VL-EHPF task and innovatively utilize Exo demonstrations to compensate for partial and dynamic Ego cues.

\subsection{Human/Robot Pose Prediction}

Early works mainly estimate human poses from Exo view \cite{sun2019deep,xiao2018simple,xu2022vitpose,maji2022yolo,guo2023back}. Recently, Ego pose has gained increasing attention \cite{grauman2024ego,liu2024single,prakash20243d,ohkawa2023assemblyhands,ma2025madiff,lin2025simhand,bao2023uncertainty} as it aligns better with human/robot-centric perception. Some works \cite{akada2025bring,wang2023scene,ohkawa2023assemblyhands,prakash20243d,liu2024single} aim to estimate the current hand poses from Ego images or videos. Meanwhile, other studies \cite{liu2022joint,bao2023uncertainty,ma2025madiff,hatano2025invisible} aim to predict future pose intentions. USST \cite{bao2023uncertainty} adopts a state space model to predict 3D hand trajectories. MADiff \cite{ma2025madiff} uses the Diffusion-based \cite{ho2020denoising} strategy to model ego motion. Despite these achievements, they mainly rely on a single visual modality for human pose estimation or coarse trajectory prediction. 

To achieve the leap from perception to execution, Visual Robot Manipulation (VRM) \cite{lynch2020learning,lynch2020language,mees2022matters,wuunleashing,yang2025ar} and Vision-Language-Action (VLA) \cite{brohan2022rt,zitkovich2023rt,yang2025egovla,kim2024openvla} have become trending topics. GCBC \cite{lynch2020learning} first learns actions from different visual conditions. MCIL \cite{lynch2020language} integrates visual and language modalities as conditions, and HULC \cite{mees2022matters} proposes a hierarchical language architecture for robust execution. Due to the limited scale of robot data, numerous studies \cite{wuunleashing,yang2025ar,yang2025egovla,yoshida2025developing} attempt to incorporate readily available human data, which has yielded remarkable progress. In this paper, we use multimodal cues to forecast fine-grained 3D hand poses in the Ego view and incorporate holistic and stable Exo demonstrations, which facilitate bridging the gap between human and robot embodiments.

\section{Method}

\subsection{Task Definition and Method Overview}

The Vision-Language guided Egocentric 3D Hand Pose Forecasting (VL-EHPF) task aims to predict the future 3D keypoints of the interacting hand(s) from the Ego perspective based on multimodal inputs. In detail, at any timestamp $t$, the model $\mathcal{M}$ takes a sequence of Ego observation frames ${{O}_{t-{T}'+1:t}}=\{{{o}_{t-{T}'+1}},{{o}_{t-{T}'+2}},\cdots {{o}_{t}}\}\in {{\mathbb{R}}^{{T}'\times H\times W\times 3}}$, a language instruction $l$, and hand pose states ${{S}_{t-{T}'+1:t}}=\{{{s}_{t-{T}'+1}},{{s}_{t-{T}'+2}},\cdots {{s}_{t}}\}\in {{\mathbb{R}}^{{T}'\times 42\times 3}}$ as multimodal inputs to forecast the 3D hand keypoints ${{\hat{S}}_{t+1:t+\bar{T}}}=\{{{\hat{s}}_{t+1}},{{\hat{s}}_{t+2}},\cdots {{\hat{s}}_{t+\bar{T}}}\}\in {{\mathbb{R}}^{\bar{T}\times 42\times 3}}$ in the future $\bar{T}$ steps. Note that 42 denotes the total number of hand joints for two hands (21 keypoints per hand), and 3 represents the 3D coordinates of each joint. This process is formulated as follows:
\begin{equation}
{{\hat{S}}_{t+1:t+\bar{T}}}=\mathcal{M}(l,{{O}_{t-{T}'+1:t}},{{S}_{t-{T}'+1:t}})
\end{equation}
where $\bar{T}$ denotes the length of future predictions, and ${T}'$ indicates the length of the observation window, which ranges between 1 and ${{T}'_{\max }}$. Practically, for cases with only a single interacting hand or missing joints, we also predict joint validity ${{\hat{V}}_{t+1:t+\bar{T}}}\in {{\{0,1\}}^{\bar{T}\times 42}}$ along with the forecasted 3D hand poses ${{\hat{S}}_{t+1:t+\bar{T}}}$.

To address the VL-EHPF task, we propose a new Exo2EgoPose framework, which is illustrated in Fig. \ref{fig:2}. First, we adopt modality-specific encoders to extract features for the input language instruction $l$, Ego observation frames ${{O}_{t-{T}'+1:t}}$, and hand pose states ${{S}_{t-{T}'+1:t}}$ at timestamp $t$. Moreover, we organize the inputs by performing multimodal tokenization and initializing distinct kinds of queries, which are fed into a Transformer-based model with causal attention for the subsequent prediction. Then, we introduce a Dual-level Exocentric Reconstruction Module (DERM), which incorporates the extra paired Exo video as supervision during training. It aims to reconstruct the Exo representations at both video and chunked frame levels, endowing the Ego-Exo cross-view capability, which facilitates modeling spatial contexts and temporal dynamics. Next, we develop a Global-to-Local Modulation Module (GLMM). It takes the reconstructed different-level Exo representations as guidance, and adopts attention mechanisms and adaptive modulation units (AMU) to refine the learned Ego features in a global-to-local manner for accurate Ego 3D hand pose forecasting. Finally, an MLP layer is employed for the final prediction of 3D hand poses.

\subsection{Multimodal Tokenization}

Given the language instruction $l$, Ego observation frames ${{O}_{t-{T}'+1:t}}$, and hand pose states ${{S}_{t-{T}'+1:t}}$, we first adopt modality-specific encoders to extract features, which are then tokenized into a unified representation space for the following reasoning and forecasting. We denote the text encoder and pose encoder as ${{\mathcal{E}}_{L}}(\cdot )$ and ${{\mathcal{E}}_{P}}(\cdot )$. For visual inputs, we employ an MAE \cite{he2022masked} encoder ${{\mathcal{E}}_{M}}(\cdot )$ and a DINOv2 \cite{oquab2023dinov2} encoder ${{\mathcal{E}}_{D}}(\cdot )$ to obtain pixel- and semantic-level features, respectively. The feature extraction process is as follows:
\begin{equation}
{{f}_{L}}={{\mathcal{E}}_{L}}(l)
\end{equation}
\begin{equation}
{{F}_{M}}=\{{{F}_{M,i}}\}_{i=t-{T}'+1}^{t}={{\mathcal{E}}_{M}}({{O}_{t-{T}'+1:t}})
\end{equation}
\begin{equation}
{{F}_{D}}=\{{{F}_{D,i}}\}_{i=t-{T}'+1}^{t}={{\mathcal{E}}_{D}}({{O}_{t-{T}'+1:t}})
\end{equation}
\begin{equation}
{{F}_{P}}=\{{{f}_{P,i}}\}_{i=t-{T}'+1}^{t}={{\mathcal{E}}_{P}}({{S}_{t-{T}'+1:t}})
\end{equation}
where ${{F}_{M,i}}= \{f_{M,i}^{\text{CLS}},f_{M,i}^{1},\cdots,f_{M,i}^{{{N}_{M}}}\}\in {{\mathbb{R}}^{({{N}_{M}}+1)\times {{C}_{v}}}}$ (similar for DINOv2 features ${{F}_{D,i}}$), and ${{f}_{P,i}}\in {{\mathbb{R}}^{{{C}_{p}}}}$. Next, we tokenize the extracted multimodal features via multi-layer perceptron (MLP) layers:
\begin{equation}
w=\text{MLP}({{f}_{L}}) \quad {{p}_{t}}=\text{MLP}({{f}_{P,t}})
\end{equation}
\begin{equation}
{{M}_{t}}=\left\{ m_{t}^{\text{CLS}},m_{t}^{1},\cdots ,m_{t}^{N} \right\}=\text{MLP}(\text{PR}({{F}_{M,t}}))
\end{equation}
\begin{equation}
{{D}_{t}}=\left\{ d_{t}^{\text{CLS}},d_{t}^{1},\cdots ,d_{t}^{N} \right\}=\text{MLP}(\text{PR}({{F}_{D,t}}))
\end{equation}
where PR means a perceiver resampler to downsample the visual token number to $N$, and all tokens are projected into the hidden size dimension $C$. Finally, we add time embeddings $e_t$ along the temporal dimension to preserve the sequential order of the inputs.

\subsection{Dual-level Exocentric Reconstruction}

Ego videos typically focus on a limited field-of-view in front of the subject with highly dynamic Ego motions. Inspired by previous Ego-Exo work \cite{grauman2024ego,huang2024egoexolearn,shi2025unsupervised,shi2026test}, we introduce the holistic and stable Exo demonstration that is paired with the given Ego observation video as supervision during training, serving as essential guidance to compensate for partial and dynamic Ego cues. To this end, we first develop a novel Dual-level Exocentric Reconstruction Module (DERM) as shown at the bottom in Fig. \ref{fig:2}. It aims to reconstruct video-level and chunked frame-level Exo representations based on the multimodal Ego inputs to construct the Ego-Exo cross-view correspondence, which facilitates comprehensively modeling the complex spatial contexts and temporal dynamics in the Ego view.

After extracting the tuple $\left\{w;{{M}_{t-{T}'+1:t}};{{D}_{t-{T}'+1:t}};{{p}_{t-{T}'+1:t}} \right\}$ of multimodal tokens, we further randomly initialize a sequence of learnable queries $Q=\{{{q}_{p}};{{q}_{v}};{{q}_{f}};{{q}_{e}}\}$, which serves as the placeholder for the Transformer prediction. Specifically, ${{q}_{p}}\in {{\mathbb{R}}^{\bar{T}\times C}}$ denotes pose queries, ${{q}_{v}}\in {{\mathbb{R}}^{1\times C}}$ represents the video-level Exo query, and ${{q}_{f}}\in {{\mathbb{R}}^{\bar{T}\times C}}$ means frame-level Exo queries. In addition, following the pioneering method GR-1 \cite{wuunleashing}, we also incorporate Ego queries ${{q}_{e}}\in {{\mathbb{R}}^{(N+1)\times C}}$ for the Ego frame reconstruction. The above tokens and queries are then fed into a multimodal Transformer architecture $\mathcal{T}(\cdot )$, which can be formulated as follows:
\begin{equation}
\begin{aligned}
\left\{q'_{p}, q'_{v}, q'_{f}, q'_{e}\right\}
= \mathcal{T}\big(
  &\big\{w, M_{t-T'+1}, D_{t-T'+1}, P_{t-T'+1},\cdots, \\
  &\cdots,w, M_{t}, D_{t}, P_{t}\big\},\, q_{p}, q_{v}, q_{f}, q_{e}
\big)
\end{aligned}
\end{equation}
where ${{q}'_{p}}$, ${{q}'_{v}}$, ${{q}'_{f}}$, and ${{q}'_{e}}$ denote the decoded query sequences.

Furthermore, we incorporate an untrimmed exocentric (Exo) video paired with the current Ego observation video $O$. It is recorded by a stable camera and provides holistic global scene contexts. The Exo observation video denoted as ${{O}^{exo}}=\{o_{1}^{exo},o_{2}^{exo},\cdots o_{{{T}_{vid}}}^{exo}\}$, contains the entire activity process of the episode with $T_{vid}$ frames in total. Then, we define dual Exo demonstrations at the video-level and chunked frame-level. In detail, We select the Exo video clip aligned with the past observations and subsequent forecasting window of the Ego video (i.e. range from timestamp $t-{T}'+1$ to $t+\bar{T}$) as the video-level Exo demonstration, denoted as $O_{t-{T}'+1:t+\bar{T}}^{exo}$. We employ the frozen MAE encoder ${{\mathcal{E}}_{M}}(\cdot )$ and the perceiver resampler (PR) to extract their features, and average the $\text{[CLS]}$ token of each frame to obtain the video-level Exo representation, as follows:
\begin{equation}
{{V}_{t-{T}'+1:t+\bar{T}}}=\left \{v_{i}^{\text{CLS}},v_{i}^{1},\cdots v_{i}^{N}\right \}_{i=t-{T}'+1}^{t+\bar{T}}=\text{PR}({{\mathcal{E}}_{M}}(O_{t-{T}'+1:t+\bar{T}}^{exo}))
\end{equation}
\begin{equation}
{{v}^{exo}}=\frac{1}{{T}'+\bar{T}}\sum\limits_{i=t-{T}'+1}^{t+\bar{T}}{v_{i}^{\text{CLS}}}
\end{equation}
where ${{v}^{exo}}\in {{\mathbb{R}}^{1\times {{C}_{v}}}}$ is the video-level Exo representation, which indicates the temporal motion dynamics across the entire activity.

In addition, we also construct chunked frame-level Exo guidance $O_{t+1:t+\bar{T}}^{exo}$ by obtaining several Exo frames corresponding to the chunked sequence to be forecasted in Ego  (i.e. range from timestamp $t+1$ to $t+\bar{T}$). Note that for cases where paired Ego and Exo videos are asynchronous such as in the EgoMe \cite{qiu2025egome} dataset, we calculate the relative temporal position of each Ego clip within the full video and conduct linear alignment to determine the corresponding timestamp range in the Exo video. We extract the $\text{[CLS]}$ token of each frame within the chunk as the frame-level Exo representations denoted as ${{f}^{exo}}=\{v_{i}^{\text{CLS}}\}_{i=t+1}^{{\bar{T}}}\in {{\mathbb{R}}^{\bar{T}\times {{C}_{v}}}}$, which mainly concentrate on the detailed spatial contextual information within the chunked frames in the future. Next, we project the decoded video-level and frame-level queries ${{q}'_{v}}$ and ${{q}'_{f}}$ into the dimension $C_v$ consistent with the Exo representations, and incorporate video-level and chunked frame-level reconstruction losses $L_{V}$ and $L_{F}$ for comprehensive Ego-to-Exo reconstruction, which can be formulated as follows:
\begin{equation}
{{q}''_{v}}=\text{MLP}({{q}'_{v}}) \quad {{q}''_{f}}=\text{MLP}({{q}'_{f}})
\end{equation}
\begin{equation}
{{L}_{V}}=\frac{1}{{{C}_{v}}}\sum\limits_{d=1}^{{{C}_{v}}}{{{({{{{q}}}''_{v}}(d)-{{v}^{exo}}(d))}^{2}}}
\end{equation}
\begin{equation}
{{L}_{F}}=\frac{1}{{\bar{T}}}\frac{1}{{{C}_{v}}}\sum\limits_{i=1}^{{\bar{T}}}{\sum\limits_{d=1}^{{{C}_{v}}}{{{({{{{q}}}''_{f,i}}(d)-f_{i}^{exo}(d))}^{2}}}}
\end{equation}
The above reconstruction losses effectively constrain the decoded queries from the Ego multimodal inputs to align with the Exo video-level and chunked frame-level representations, which derive from the paired holistic and stable Exo demonstration video, facilitating modeling spatial contexts and temporal dynamics in Ego scenarios. In addition, we perform Ego frame reconstruction. Specifically, we feed the decoded Ego queries into a multi-block MAE-style decoder $\mathcal{E}_E$ to reconstruct the future third Ego frame relative to the current timestep, and the corresponding loss is denoted as $L_E$. The overall reconstruction loss can be formulated as follows:
\begin{equation}
{{L}_{recon}}={{\lambda }_{V}}{{L}_{V}}+{{\lambda }_{F}}{{L}_{F}}+{{\lambda }_{E}}{{L}_{E}}
\end{equation}
where $\lambda_{V}$, $\lambda_{F}$, and $\lambda_{E}$ are coefficients for balancing multiple losses.

\subsection{Global-to-Local Modulation Module}

\begin{figure}[!t]
\centering
\includegraphics[width=0.94\linewidth]{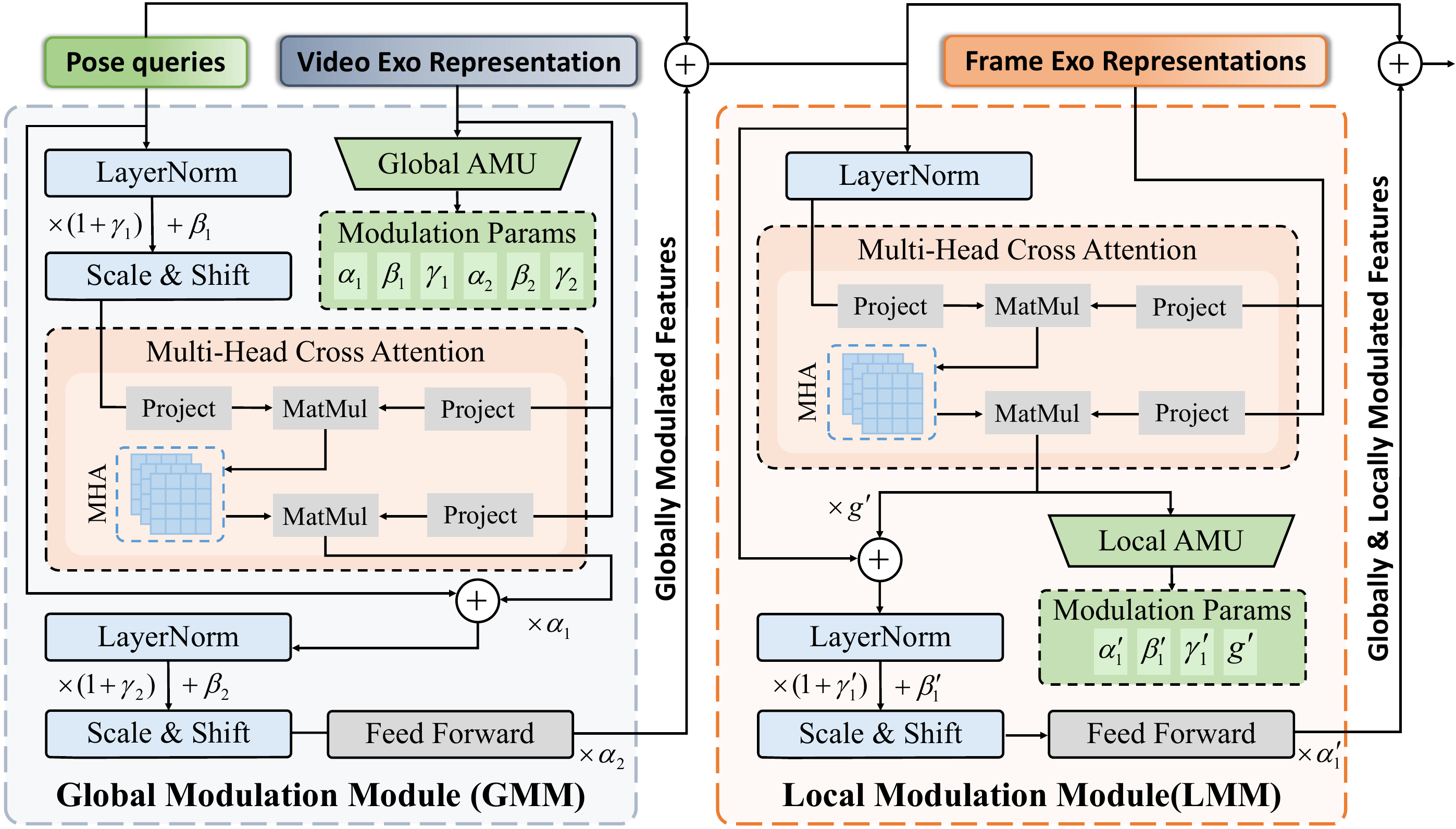}
\caption{Illustration of the Global-to-Local Modulation Module (GLMM). Given the reconstructed Exo representations, it performs feature refinement from global to local levels via cross attention and adaptive modulation units (AMU).}
\label{fig:3}
\end{figure}

After reconstructing the dual-level Exo demonstrations, we develop a Global-to-Local Modulation Module (GLMM) as shown in Fig. \ref{fig:3} to comprehensively inject the different-level Exo-view guidance into the learned Ego pose features. In detail, GLMM utilizes the reconstructed hierarchical Exo representations to progressively refine the learned Ego features in a global-to-local manner, and employs attention mechanisms and adaptive modulation units (AMU) for feature alignment and distribution calibration, respectively.

In the aforementioned DERM, we obtain the decoded pose queries ${{{q}}'_{p}}\in {{\mathbb{R}}^{\bar{T}\times C}}$ and the reconstructed video- and chunked frame-level Exo representations denoted as ${{q}''_{v}}\in {{\mathbb{R}}^{1\times {{C}_{v}}}}$ and ${{q}''_{f}}\in {{\mathbb{R}}^{\bar{T}\times {{C}_{v}}}}$, respectively. First, we treat ${{q}''_{v}}$, which indicates the video-level temporal motion patterns of the entire activity as the global Exo guidance. ${{q}''_{f}}$ contains the detailed frame-wise spatial contextual information that serves as the local Exo guidance. Then, we feed the pose queries ${{{q}}'_{p}}$ and global Exo guidance ${{q}''_{v}}$ into a Global Modulation Module (GMM). Because ${{q}''_{v}}$ is the aggregated representation across all frames in the video-level Exo demonstration, GMM directly computes the modulation parameters based on ${{q}''_{v}}$ through adaptive modulation units (AMU) to calibrate the feature distribution via adaptive layer normalization (AdaLN) \cite{peebles2023scalable}. Additionally, we also insert a multi-head cross-attention block into the AdaLN layers to enable the learned Ego pose features to attend to the global Exo guidance, thereby capturing the long-range temporal dynamics of the complete activity. This process can be formulated as follows:
\begin{equation}
\left\{{{\alpha }_{1}},{{\beta }_{1}},{{\gamma }_{1}},{{\alpha }_{2}},{{\beta }_{2}},{{\gamma }_{2}}\right\}=\text{GlobalAMU}({{q}''_{v}})=\text{MLP}(\text{SiLU}({{q}''_{v}}))
\end{equation}
\begin{equation}
{{q}^{g}_{m}}=\text{AdaLN}({{q}'_{p}},{{\gamma }_{1}},{{\beta }_{1}})=(1+{{\gamma }_{1}})\odot\text{LN}({{q}'_{p}}) +{{\beta }_{1}}
\end{equation}
\begin{equation}
{{\mathcal{A}}_{g}}=\alpha_1 \odot \left(\text{Softmax}(\frac{({{q}^{g}_{m}}W_{q}^{g}){{({{q}''_{v}}W_{k}^{g})}^{\top}}}{\sqrt{{{d}_{k}}}}){{q}''_{v}}W{{_{v}^{g}}}\right)+{{q}'_{p}}
\end{equation}
\begin{equation}
{{\bar{q}}^{g}_{m}}={{\alpha }_{2}}\odot \text{MLP}(\text{AdaLN}({{\mathcal{A}}_{g}},{{\gamma }_{2}},{{\beta }_{2}}))
\end{equation}
where $\alpha_1, \beta_1, \gamma_1, \alpha_2, \beta_2, \gamma_2 \in \mathbb{R}^{C}$ denote six modulation vectors, $\odot$ means Hadamard product, $W_{q}^{g}\in {{\mathbb{R}}^{C\times C}}$, $W_{k}^{g}\in {{\mathbb{R}}^{C_v\times C}}$, $W_{v}^{g}\in {{\mathbb{R}}^{C_v\times C}}$ denote learnable parameters in the attention block, $d_k$ is the hidden size, and ${{\bar{q}}^{g}_{m}}\in {{\mathbb{R}}^{\bar{T}\times C}}$ represents the globally modulated features. 

Next, we input the globally modulated features ${{\bar{q}}^{g}_{m}}$ and local Exo guidance ${{q}''_{f}}$ into a Local Modulation Module (LMM), which exhibits a structure distinct from the above GMM. Specifically, since ${{q}''_{f}}$ denotes Exo features of local chunked frames whose contents may not temporally synchronize with the learned features ${{\bar{q}}^{g}_{m}}$, the LMM first conducts multi-head cross-attention with gated residual connection for a frame-level Ego-Exo soft feature alignment, followed by the local AMU to compute the modulation parameters $\{{g}', {{\alpha }'_{1}},{{\beta }'_{1}},{{\gamma }'_{1}}\}$, which can be formulated as follows:
\begin{equation}
{{\mathcal{A}}_{l}}=\text{Softmax}\left(\frac{(\text{LN}(\bar{q}_{m}^{g})W_{q}^{l}){{({{q}''_{f}}W_{k}^{l})}^{\top}}}{\sqrt{{{d}_{k}}}}\right){{q}''_{f}}W_{v}^{l}
\end{equation}
\begin{equation}
{g}'=\text{Sigmoid}(\text{MLP}({{\mathcal{A}}_{l}})) \quad \left\{ {{\alpha }'_{1}},{{\beta }'_{1}},{{\gamma }'_{1}} \right\}=\text{MLP}(\text{SiLU}({\mathcal{A}}_{l}))
\end{equation}
\begin{equation}
\bar{q}_{m}^{l}={{{\alpha }}'_{1}}\odot \text{MLP}(\text{AdaLN}({g}'\odot {{\mathcal{A}}_{l}}+\bar{q}_{m}^{g},{{{\gamma }}'_{1}},{{{\beta }}'_{1}}))
\end{equation}
where $g'\in \mathbb{R}^{C}$ denotes the learnable gate parameter and ${{\bar{q}}^{l}_{m}}\in {{\mathbb{R}}^{\bar{T}\times C}}$ represents the final globally and locally modulated features.

Finally, the future Ego 3D hand poses ${{\hat{S}}_{t+1:t+\bar{T}}}$ and joint validity masks ${{\hat{V}}_{t+1:t+\bar{T}}}$ are jointly predicted by an MLP layer and are supervised by standard smoothL1 and BCE losses, which are denoted as $L_P$ and $L_{va}$, respectively. The overall loss is as follows:
\begin{equation}
{{L}_{overall}}={{L}_{recon}}+{{\lambda }_{P}}{{L}_{P}}+{{\lambda }_{va}}{{L}_{va}}
\end{equation}
where ${{\lambda }_{P}}$ and ${{\lambda }_{va}}$ are coefficients for balancing loss items.

\section{Experiments}

\subsection{Experimental Settings}

\begin{table*}[!t]
\centering
\caption{Quantitative results on the \textit{test} set of the \textit{AssemblyHands}, \textit{Ego-Exo4D}, and novel \textit{EgoMe-pose} benchmarks.}
\label{tab:1}
\scalebox{0.91}{
\begin{tabular}{p{2.8cm}<{\raggedright}|p{1.5cm}<{\centering}p{1.5cm}<{\centering}|p{1.5cm}<{\centering}p{1.5cm}<{\centering}|p{1.5cm}<{\centering}p{1.5cm}<{\centering}}
\toprule
\multirow{2}{*}{Methods} & \multicolumn{2}{c|}{\textit{AssemblyHands (15 fps)}}  & \multicolumn{2}{c|}{\textit{Ego-Exo4D (30 fps)}}   & \multicolumn{2}{c}{\textit{EgoMe-pose (15 fps)}}   \\    \cmidrule{2-7} 
& MPJPE $\downarrow$ & MPJVE $\downarrow$ & MPJPE $\downarrow$ & MPJVE $\downarrow$ & MPJPE $\downarrow$ & MPJVE $\downarrow$ \\ \midrule  
\rowcolor{gray!15}Random Forecasting  & 745.35 & 383.80 & 736.80 & 393.63 & 794.00 & 397.77 \\
USST \cite{bao2023uncertainty} & 52.58 & 19.75 & 63.07 & 22.99 & 71.93 & 73.67 \\
GCBC \cite{lynch2020learning} & 45.38 & 12.13 & 49.97 & 21.86 & 63.43 & 71.45 \\
MCIL \cite{lynch2020language} & 42.58 & 11.05 & 47.17 & 17.53 & 62.81 & 70.48 \\
HULC \cite{mees2022matters} & 41.88 & 9.75 & 46.73 & 17.34 & 60.81 & 69.83 \\
GR-1 \cite{wuunleashing} & 40.03 & 8.69 & 44.29 & 16.44 & 59.10 & 69.03 \\
AR-VRM \cite{yang2025ar} & 33.39 & 7.86 & 44.46 & 16.33 & 56.06 & 67.66 \\
Our Exo2EgoPose & \textbf{25.83} & \textbf{6.35} & \textbf{36.44} & \textbf{16.06} & \textbf{49.44} & \textbf{61.03} \\
\bottomrule
\end{tabular}}
\end{table*}

\subsubsection{Benchmarks}

We evaluate our method on the existing \textit{AssemblyHands} and \textit{Ego-Exo4D} benchmarks, and further construct a brand-new \textit{EgoMe-pose} benchmark for a comprehensive evaluation.

\noindent \textbf{\textit{AssemblyHands}} \cite{ohkawa2023assemblyhands} provides 3D hand annotations built upon \textit{Assembly101} \cite{sener2022assembly101}. It contains synchronized Ego-Exo videos and provides 3.0M frames with 3D hand pose annotations. Moreover, we correlate the annotated video clips with the textual action annotations in \textit{Assembly101} to match our VL-EHPF task. Finally, we curate \textit{6120}/\textit{355}/\textit{355} activity episodes for \textit{train}/\textit{val}/\textit{test} splits, respectively.

\noindent \textbf{\textit{Ego-Exo4D}} \cite{grauman2024ego} is one of the largest datasets. We use the \textit{Ego-Exo4D EgoPose} benchmark, which contains 68K manual and 4.3M automatic annotations. In addition, we correlate hand poses to atomic action descriptions to construct multimodal triplets. Finally, we curate \textit{10869}/\textit{1540}/\textit{1540} episodes for the \textit{train}/\textit{val}/\textit{test} splits.

\noindent \textbf{\textit{EgoMe-pose}} is a brand-new benchmark based on the recent \textit{EgoMe} \cite{qiu2025egome} dataset. \textit{EgoMe} contains paired but asynchronous Exo and Ego videos with detailed textual and temporal annotations. Since the pose annotation is not available, we apply an automatic labeling and filtering pipeline to construct the \textit{EgoMe-pose} benchmark:

\textbf{(1) Data selection}: We retain the correctly following Ego-Exo video pairs and exclude false imitation cases in the \textit{EgoMe} dataset.

\textbf{(2) Hand detection}: Before 3D hand pose labeling, we first utilize an on-the-shelf hand detector with a threshold ${{\sigma }_{d}}\ge 0.5$ to extract bounding boxes of hand(s) in the video frames.

\textbf{(3) Hand pose labeling}: We employ InterHand \cite{moon2020interhand2} to estimate the relative 3D hand poses. Then, we adopt RootNet \cite{Moon_2019_ICCV_3DMPPE} to predict the absolute depth of the hand roots. Furthermore, we perform the relative-to-absolute affine transformation of the coordinate system.

\textbf{(4) Annotation filtering}: To ensure the accuracy of automatic labeling, we filter the obtained pose annotations. In detail, we only retain cases where the hand confidence score ${{\sigma }_{d}}\ge 0.6$ and more than 95$\%$ of frames in the episode are validly annotated. 

We curate \textit{6067}/\textit{1344}/\textit{2648} episodes for the \textit{train}/\textit{val}/\textit{test} splits.

\subsubsection{Evaluation Metrics}

Following common settings \cite{hatano2025invisible,chen2025egoagent}, we adopt Mean Per Joint Position Error (MPJPE) and Mean Per Joint Velocity Error (MPJVE) in millimeters (mm). MPJPE computes the Euclidean Distance (ED) between root-aligned predicted and ground-truth joints to evaluate position accuracy, while MPJVE computes ED between their joint displacements to assess hand motion.

\subsubsection{Implementation Details}

We use the text branch of a pretrained CLIP (ViT-B/32) \cite{radford2021learning} as the text encoder, and adopt dual encoders (i.e., pretrained MAE \cite{he2022masked} and DINOv2 \cite{oquab2023dinov2}) for the input frames. For hand pose states, we adopt a trainable HandFormer \cite{shamil2024utility} as the pose encoder. In addition, we use the randomly initialized 12-layer Transformer with causal attention \cite{radford2019language} as our multimodal Transformer architecture, with dropout set to 0.1. The feature dimensions ${{C}_{v}}$, ${{C}_{l}}$, and ${{C}_{p}}$ are 768, 512, and 384, respectively. The hidden size $C$ is set to 384. The sample rates of videos and hand poses are 15 fps for \textit{AssemblyHand} and \textit{EgoMe-pose}, and 30 fps for \textit{Ego-Exo4D}, and the visual frames are resized to 224 $\times$ 224. The number of visual tokens is ${{N}_{M}}$=${{N}_{D}}$=$196$, which are resampled to $N$=$9$. The sequence lengths ${{T}'_{\max }}$ and $\bar{T}$ are both set to 10. The balance coefficients ${{\lambda }_{P}}$, ${{\lambda }_{va}}$, ${{\lambda }_{E}}$, ${{\lambda }_{V}}$, and ${{\lambda }_{F}}$ are set to 3e3, 1.0, 1.0, 1.0, and 0.1, respectively. For training, the parameters are optimized by AdamW with a weight decay of 1e-4, and the batch size is set to 48. The epoch number is set to 20, with a warmup epoch, and we apply the cosine learning rate decay strategy from 1e-4 to 5e-6.

\subsection{Comparison with State-of-the-Art Methods}

Since the VL-EHPF task focuses on inferring fine-grained actions based on multimodal inputs, it aligns better with the VRM/VLA paradigm rather than unimodal coarse-level Ego motion prediction. Therefore, we mainly compare our Exo2EgoPose with VRM/VLA methods (i.e., GCBC \cite{lynch2020learning}, MCIL \cite{lynch2020language}, HULC \cite{mees2022matters}, GR-1 \cite{wuunleashing}, AR-VRM \cite{yang2025ar}), and include one representative 3D hand trajectory forecasting method (i.e., USST \cite{bao2023uncertainty}) for a comprehensive comparison in Table \ref{tab:1}. We re-implement these methods and apply the same backbones \cite{radford2021learning,he2022masked,oquab2023dinov2,shamil2024utility,radford2019language} as in our method for a fair comparison.

We evaluate the above methods on \textit{AssemblyHands}, \textit{Ego-Exo4D}, and our newly constructed \textit{EgoMe-pose} benchmarks in Table \ref{tab:1}. Compared with the random forecasting baseline in the first row, all other methods achieve much lower error values, showing their applicability for the VL-EHPF task. Notably, our Exo2EgoPose framework significantly outperforms other related methods on all benchmarks. Specifically, compared to USST \cite{bao2023uncertainty}, GCBC \cite{lynch2020learning}, MCIL \cite{lynch2020language}, and HULC \cite{mees2022matters} methods, our approach yields remarkable reductions of over 10 MPJPE. GR-1 \cite{wuunleashing} is a pioneering method that incorporates large-scale videos in the Ego4D \cite{grauman2022ego4d} dataset for generative pretraining. However, our method decreases the MPJPE/MPJVE by 14.20/2.34 on \textit{AssemblyHands}, 7.85/0.38 on \textit{Ego-Exo4D}, and 9.66/8.00 on \textit{EgoMe-pose} compared to GR-1. In addition, we also extend AR-VRM \cite{yang2025ar} by incorporating Exo data for analogical reasoning, which achieves high performance. Nevertheless, Exo2EgoPose surpasses it by 7.56 MPJPE and 1.51 MPJVE on \textit{AssemblyHands}, by 8.02 MPJPE and 0.27 MPJVE on \textit{Ego-Exo4D}, and by 6.62 MPJPE and 6.63 MPJVE on our \textit{EgoMe-pose}. The above results demonstrate the effectiveness of our Exo2EgoPose, which performs dual-level Exo reconstruction and modulates the learned Ego features from global to local levels.

\subsection{Ablation Study}

\begin{table}[!t]
\centering
\caption{Ablation results on the \textit{val} set of the \textit{AssemblyHands} dataset. ``VER" and ``CFER" denote the video-level and chunked frame-level Exo reconstruction, respectively. ``T" denotes replacing GMM and LMM with vanilla Transformer layers.}
\label{tab:2}
\scalebox{0.83}{
\begin{tabular}{p{0.8cm}<{\centering}p{0.8cm}<{\centering}p{0.8cm}<{\centering}p{0.8cm}<{\centering}|p{1.0cm}<{\centering}p{1.0cm}<{\centering}}
\toprule
\multicolumn{1}{c}{GMM} & \multicolumn{1}{c}{LMM} & \multicolumn{1}{c}{VER} & \multicolumn{1}{c|}{CFER} & \multicolumn{1}{c}{MPJPE $\downarrow$} & \multicolumn{1}{c}{MPJVE $\downarrow$} \\
\midrule  
\checkmark  & \checkmark  & \checkmark   & \checkmark   & \textbf{25.37}  & \textbf{6.29}  \\
T  & T & \checkmark   & \checkmark   & 26.44  & 6.57 \\
  & \checkmark  & \checkmark   & \checkmark   & 27.03 & 6.56  \\
  \checkmark  &   & \checkmark   & \checkmark   & 27.26 &  6.48 \\
   &   & \checkmark   & \checkmark   & 30.11 & 7.76\\
\midrule
   &   &    & \checkmark   &  32.94  & 8.07 \\
   &   &  \checkmark  &   & 33.22  &  8.06  \\
   &   &    &   & 39.70  &  8.87 \\
\bottomrule
\end{tabular}}
\end{table}

\begin{figure*}[!t]
\centering
\includegraphics[width=0.90\linewidth]{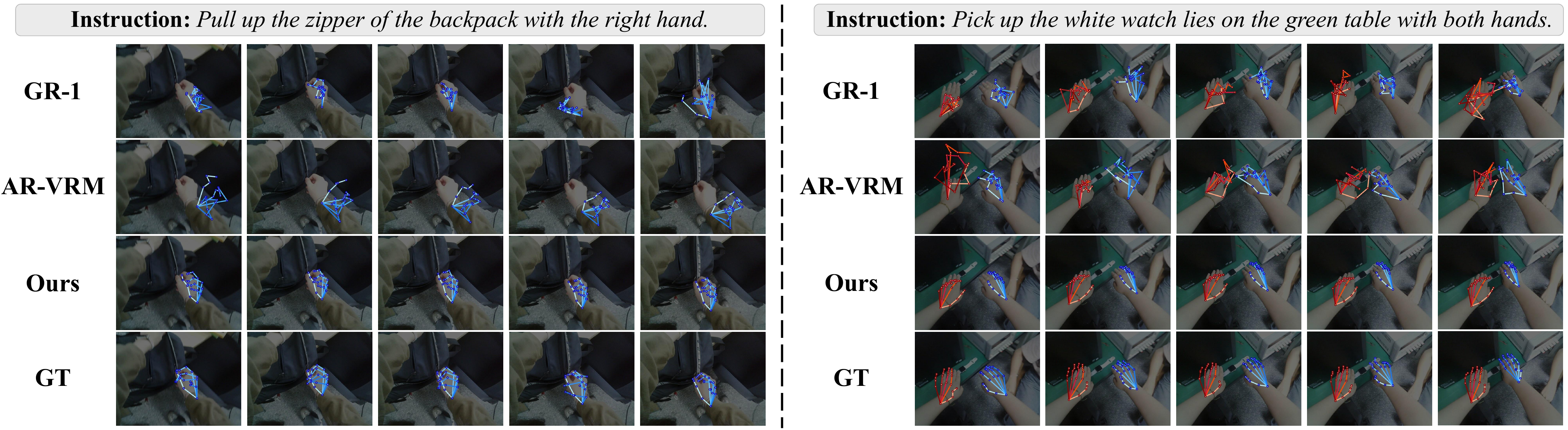}
\caption{Results of the forecasted Ego 3D hand poses of our Exo2EgoPose and comparison methods (downsampled for brevity).}
\label{fig:4}
\end{figure*}

In Table \ref{tab:2}, we conduct ablation experiments to analyze the effectiveness of the key components in our framework. In the second row, we replace the proposed GMM and LMM with vanilla Transformer layers, causing a rise of 1.07 in MPJPE and an increase of 0.28 in MPJVE, which shows the superiority of the designed GMM and LMM based on cross-attention and adaptive modulation. In the third row, we remove the Global Modulation Module (GMM), exhibiting performance drops of 1.66 MPJPE and 0.27 MPJVE. We also remove the Local Modulation Module (LMM) in the fourth row, which leads to 1.89 MPJPE and 0.19 MPJVE degradation. The above results show the respective effectiveness of GMM and LMM, and further validate that GMM excels in capturing temporal motion trends (evidenced by MPJVE), while LMM yields greater improvements in spatial positions (evidenced by MPJPE). Then, we remove the GMM and LMM simultaneously in the fifth row, which causes a remarkable performance decline of 4.74 MPJPE and 1.47 MPJVE, showing the effectiveness of the proposed GLMM in feature alignment and distribution calibration to progressively inject the Exo guidance into the learned Ego features for accurate hand pose forecasting.

Next, we disable video-level Exo reconstruction in the sixth row, and the errors rise significantly by 2.83 MPJPE and 0.31 MPJVE compared with those in the fifth row. Meanwhile, in the seventh row, we discard the chunked frame-level Exo reconstruction, leading to a drop of 3.11 MPJPE and 0.30 MPJVE. The above results indicate the effectiveness of reconstructing Exo demonstrations at video- and chunked frame-level, even though the Exo information does not serve as guidance for feature modulation. Moreover, we completely remove Exo demonstrations in the last row, which causes severe degradation of 9.59 in MPJPE and 1.11 in MPJVE. It further emphasizes the effectiveness of Exo demonstrations, which facilitate modeling complex spatial contexts and temporal dynamics.

\subsection{In-depth Analysis}

\subsubsection{Analysis of Qualitative Results}

In Fig. \ref{fig:4}, we show two examples of the forecasted Ego 3D hand poses of our Exo2EgoPose and two recent approaches \cite{yang2025ar,wuunleashing}. Specifically, we map the 3D hand joints into 2D frames and perform linear scaling and alignment with a temporal stride of 2 for better visualization. In the left example, the subject is going to ``pull up the zipper of the backpack with the right hand". The GR-1 in the first row fails to generate plausible hand structures. In the second row, while the AR-VRM roughly models the hand structures with severe errors in hand orientation and detailed poses, indicating it fails to understand the process of ``pulling a zipper". In contrast, our Exo2EgoPose correctly comprehends this process and forecasts the fine-grained hand actions precisely. The right example is more challenging, requiring bimanual grasping of the watch on the table. Although GR-1 distinguishes the left and right hands, it still struggles to model their respective hand structures. AR-VRM shows slight improvements and forms basic hand shapes. Nevertheless, its predicted joint positions and motions still deviate significantly from the ground-truths. Finally, our method accurately predicts the hand poses, including the contraction process of the right hand during grasping. The above qualitative results further validate the effectiveness of our approach, which incorporates the Exo demonstration videos and utilizes the reconstructed Exo representations as guidance to modulate the learned Ego features.

\subsubsection{Analysis of Hyperparameter Sensitivity}

\begin{figure}[!t]
\centering
\includegraphics[width=1.0\linewidth]{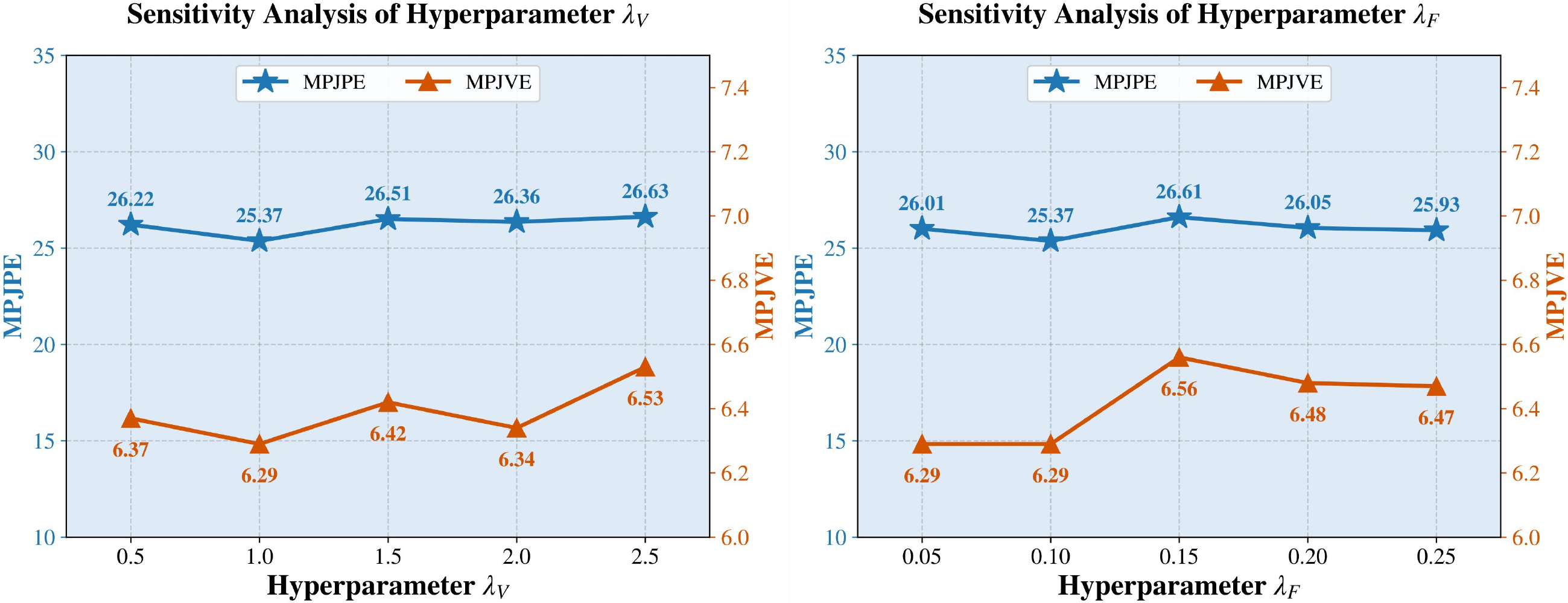}
\caption{Sensitivity analysis for hyperparameters $\lambda_V$ and $\lambda_F$.}
\label{fig:6}
\end{figure}

In Fig. \ref{fig:6}, we perform a sensitivity analysis on $\lambda_V$ and $\lambda_F$ on \textit{AssemblyHands} to evaluate the robustness of the DERM. In detail, we set $\lambda_V$ in the range of 0.5 to 2.5, while $\lambda_F$ is from 0.05 to 0.25. The results show that the model is robust to changes in $\lambda_V$ and $\lambda_F$, i.e., the MPJPE and MPJVE metrics only fluctuate within small ranges of 1.26 and 0.27, respectively.  Finally, our method yields the minimum MPJPE of 25.37 and MPJVE of 6.29 when $\lambda_V=1.0$ and $\lambda_F=0.1$. These results validate the effectiveness of our DERM, which consistently maintains high performance across a broad range of hyperparameter settings.

\subsubsection{Analysis of Feature Modulation}

\begin{figure}[!t]
\centering
\includegraphics[width=1.0\linewidth]{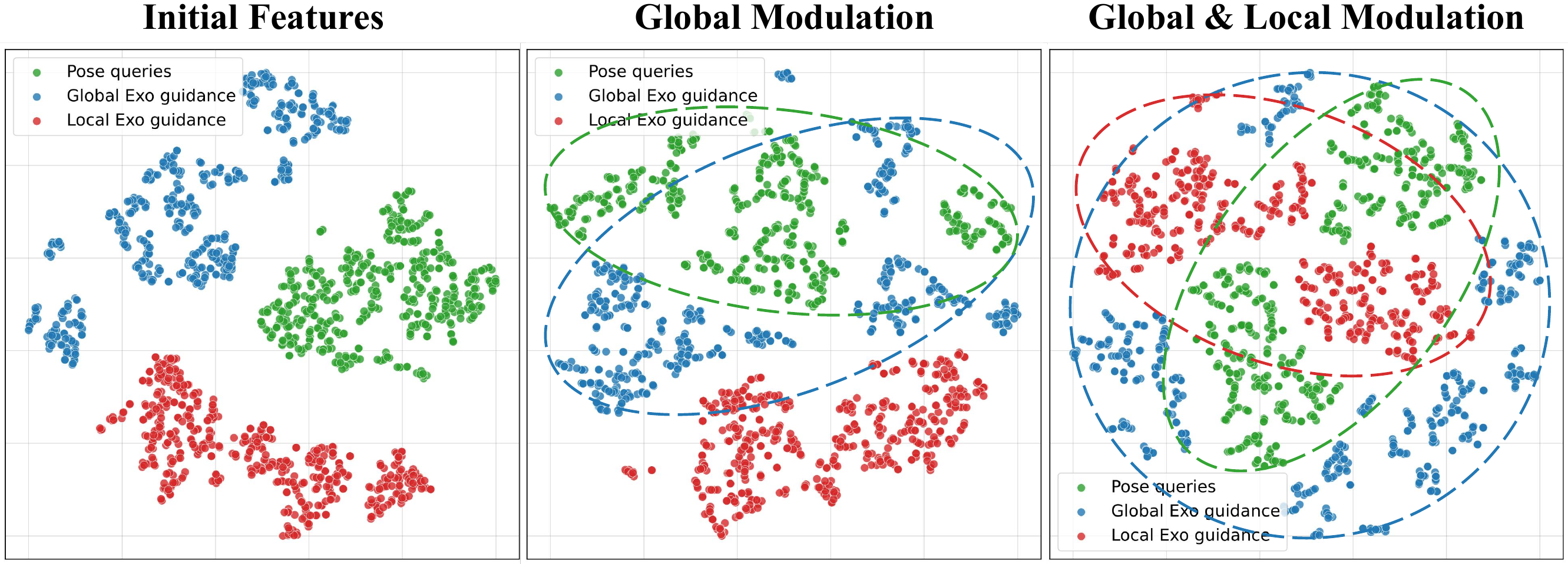}
\caption{Visualization of representation distributions for the decoded pose queries at different stages, global Exo guidance, and local Exo guidance (best viewed in color).}
\label{fig:5}
\end{figure}

To intuitively visualize the progressive global-to-local feature modulation process, we use t-SNE \cite{van2008visualizing} to project the decoded pose queries at different stages, the global- and local-level Exo representations onto 2D planes in Fig. \ref{fig:5}.

In the first column, we visualize the representations of the initial decoded pose queries, global and local Exo guidance, and the results show that the initial representations present significantly different distributions. In the second column, we visualize the pose queries after the global-level feature modulation. The results show that the feature clusters of the pose queries and the global Exo guidance are close to each other, indicating that the global-level modulation effectively adjusts the distribution of the learned pose features based on the reconstructed video-level Exo representations. Finally, in the third column, we visualize the globally and locally modulated features. The results reveal that the feature clusters of the three components are tightly fused, which demonstrates that the local-level modulation further injects fine-grained chunked frame-level Exo guidance knowledge into the globally modulated pose queries. These visualization results validate that the proposed GLMM effectively conducts feature alignment and distribution calibration for the learned Ego pose features based on the reconstructed hierarchical Exo representations, facilitating Ego 3D hand pose forecasting.

\begin{table}[!t]
\centering
\caption{The inference FLOPs and number of parameters of the baseline model and key components of our Exo2EgoPose.}
\label{tab:4}
\scalebox{0.95}{
\begin{tabular}{p{1.6cm}<{\raggedright}|p{1.0cm}<{\centering}|p{1.0cm}<{\centering}|p{1.0cm}<{\centering}|p{1.0cm}<{\centering}|p{1.0cm}<{\centering}}
\toprule
      & Baseline & GMM & LMM & VER & CFER \\ \midrule
FLOPs (G)  & 802.100 &	0.254 & 0.354 & 0.006 & 0.062  \\ \hline
$\#$Params (M)& 236.678 & 2.069 & 1.774 & 0.296 & 0.296 \\ 
\bottomrule
\end{tabular}}
\end{table}

\subsubsection{Analysis of Model Complexity}

We conduct experiments to evaluate the computational complexity and parameter overhead of the baseline model and the key components of our Exo2EgoPose in Table \ref{tab:4}. We remove all of our newly proposed modules from Exo2EgoPose as the baseline model, which has an inference computational complexity of 802.100 GFLOPs and a parameter count of 236.678 M. In contrast, the proposed key components (i.e., GMM, LMM, VER, and CFER) present remarkable efficiency. In detail, in the second and third columns, we report the computational complexity and the number of parameters of GMM and LMM, which mainly consist of MLP, AdaLN, and cross-attention layers, and only introduce 0.254 and 0.354 GFLOPs as well as 2.069 and 1.774 M parameters, respectively. In addition, for the video-level and chunked frame-level Exo reconstruction (VER and CFER), we only need to initialize extra query sequences and projection layers for Transformer reasoning during inference. Consequently, they incorporate negligible computational complexity and parameter overhead (i.e., 0.006 GFLOPs and 0.296 M parameters for VER, and 0.062 GFLOPs and 0.296 M parameters for CFER). Considering the remarkable performance gains, the additional computational and parameter overhead introduced by the key components is marginal.

\subsubsection{Analysis of Human-to-Robot Transfer}

We observe that Ego 3D hand poses can serve as a natural and explicit unified representation to bridge human action and robot manipulation. Therefore, we conduct experiments on the challenging ABC$\rightarrow$D (i.e., trained on A, B, and C scenes and evaluated on the unseen scene D) setting on the robotic \textit{CALVIN} benchmark to evaluate the human-to-robot capability of our Exo2EgoPose. In detail, we utilize a pretrained Exo2EgoPose model as a teacher network and finetune a student policy network \cite{wuunleashing} by introducing a distillation loss on the intermediate representations. As shown in Table \ref{tab:3}, our method exhibits impressive capabilities in robotic tasks. Specifically, our method achieves higher success rates in sequences of 2 to 5 tasks, showing its long-horizon task completion ability. Moreover, our method yields the highest average success length and rate, which outperforms the second-place AR-VRM \cite{yang2025ar} by 0.09 in the length of successfully completed tasks and 1.6$\%$ in success rate. The results validate that using the Exo demonstrations for accurate Ego 3D human hand pose forecasting provides highly informative and transferable representations, which effectively benefit downstream robotic manipulation tasks under zero-shot and long-horizon settings.

\begin{table}[!t]
\centering
\caption{Analysis of human-to-robot transfer capability under the ABC$\rightarrow$D setting on the \textit{CALVIN} benchmark. ``Avg.L." denotes the average length of successfully completed tasks in a sequence, ``Avg.R." represents the average success rate.}
\label{tab:3}
\scalebox{0.87}{
\begin{tabular}{p{1.8cm}<{\raggedright}|p{0.6cm}<{\centering}p{0.6cm}<{\centering}p{0.6cm}<{\centering}p{0.6cm}<{\centering}p{0.6cm}<{\centering}|p{0.65cm}<{\centering}p{0.65cm}<{\centering}}
\toprule
\multirow{2}{*}{Methods} & \multicolumn{5}{c|}{Tasks completed in a row} & \multirow{2}{*}{Avg.L.} & \multirow{2}{*}{Avg.R.} \\
& 1 & 2 & 3 & 4& 5 & &  \\ \midrule
MCIL \cite{lynch2020language} & 0.304 & 0.013 & 0.002 & 0.000 & 0.000 & 0.31 & 6.4$\%$ \\
RT-1 \cite{brohan2022rt} & 0.533 & 0.222 & 0.094 & 0.038 & 0.013 & 0.90 & 18.0$\%$ \\
HULC \cite{mees2022matters} & 0.418 & 0.165 & 0.057 & 0.019 & 0.011 & 0.67 & 13.4$\%$ \\
MT-R3M \cite{nair2022r3m} & 0.529 & 0.234 & 0.105 & 0.043 & 0.018 & 0.93 & 18.6$\%$ \\
GR-1 \cite{wuunleashing} & 0.854 & 0.712 & 0.596 & 0.497 & 0.401 & 3.06 & 61.2$\%$ \\
AR-VRM \cite{yang2025ar} & \textbf{0.901} & \underline{0.759} & \underline{0.642} & \underline{0.531} & \underline{0.461} & \underline{3.29} & \underline{65.9$\%$} \\
Exo2EgoPose & \underline{0.859} & \textbf{0.762} & \textbf{0.669} & \textbf{0.589} & \textbf{0.498} & \textbf{3.38} & \textbf{67.5$\%$} \\
\bottomrule
\end{tabular}}
\end{table}

\section{Conclusion}

In this paper, we investigate the under-explored Vision-Language guided Egocentric 3D Hand Pose Forecasting (VL-EHPF) task, which aims to forecast the 3D hand pose in the Ego view conditioned on multimodal cues (i.e., visual observations, a language instruction, and pose states). To tackle the challenges of limited field-of-view and dynamic motion, we innovatively incorporate Exo-view demonstrations and propose an Exo2EgoPose framework. It first models the spatial contexts and temporal dynamics by reconstructing Exo representations at video and frame levels via a Dual-level Exocentric Reconstruction Module (DERM). In addition, the Global-to-Local Modulation Module (GLMM) utilizes the reconstructed representations for progressive feature refinement, facilitating comprehensive Exo guidance for accurate forecasting. Extensive experiments on \textit{AssemblyHands}, \textit{Ego-Exo4D}, brand-new \textit{EgoMe-pose}, and robotic \textit{CALVIN} benchmarks demonstrate the superiority of our method.



\begin{acks}
This work was supported by the National Natural Science Foundation of China (No. U23A20286, No. 62301121), Sichuan Science and Technology Program (No.2026NSFSC1478).
\end{acks}

\bibliographystyle{ACM-Reference-Format}
\balance
\bibliography{references}


\appendix

\setcounter{table}{0}   
\setcounter{figure}{0}
\setcounter{equation}{0}
\renewcommand{\thetable}{A\arabic{table}}
\renewcommand{\thefigure}{A\arabic{figure}}
\renewcommand{\theequation}{A\arabic{equation}}

\section{Details of State-of-the-Art Methods}

In this section, we introduce some technical and re-implementation details of the state-of-the-art comparison methods (i.e., USST \cite{bao2023uncertainty}, GCBC \cite{lynch2020learning}, MCIL \cite{lynch2020language}, HULC \cite{mees2022matters}, GR-1 \cite{wuunleashing}, and AR-VRM \cite{yang2025ar}).

\subsection{USST}

USST \cite{bao2023uncertainty} is the representative work for egocentric 3D hand trajectory forecasting. It takes the Egocentric (Ego) frames and the historical hand trajectory as inputs to forecast the coarse Ego 3D hand trajectory in the future. However, it does not leverage the detailed hand pose states and the language instruction. USST introduces a state space model and an attention-based state transition module with an emission module for this task. In this paper, we first modify the forecasting head in USST to enable forecasting future 3D hand joints. Then, we re-implement this method by migrating its key components, such as uncertainty estimation and uncertainty-aware losses, and the balance coefficient of the uncertainty-aware losses is set to 3e-3. During forecasting, the model takes the Ego observation frames and prior hand states as input without the language instruction, which is consistent with the setting in the original USST. Finally, due to the absence of explicit language instruction, the re-implemented USST yields a low performance in our proposed task, which demonstrates a substantial discrepancy in task settings between the hand trajectory forecasting and our Vision-Language guided Egocentric 3D Hand Pose Forecasting (VL-EHPF) tasks.

\subsection{GCBC}

GCBC \cite{lynch2020learning} is a pioneering work for the Visual Robot Manipulation (VRM) task. It makes the first exploration to learn fine-grained robotic control latents from human teleoperated play data in a self-supervised manner. In detail, it takes play data (i.e., continuous logs of low-level visual observations and fine-grained actions) as inputs to learn a latent space, which is reused at test time to execute specific goals. To re-implement this method for our VL-EHPF task, we practically employ visual observations, a language instruction, and hand pose states as multimodal inputs, which is consistent with the common settings in the mainstream VRM/VLA works. Then, we incorporate an additional token serving as the action latent space into the autoregressive Transformer for the subsequent learning from the multimodal priors to the Ego 3D hand poses in the future.

\subsection{MCIL}

MCIL \cite{lynch2020language} is an important VRM/VLA work that makes the first exploration of incorporating free-form natural language conditioning into robot task executions. Specifically, this approach learns perception from natural language instruction, dense visual frame pixels, continuous control, and complex tasks. To re-implement MCIL, we integrate its core components (i.e., continuous latent plan, plan proposal module, and plan recognition module) based on its officially released code. Specifically, the plan recognition module is based on Bi-LSTM layers, which serve as a posterior encoder that summarizes the holistic intention during training. In contrast, the plan proposal module serves as a prior network, predicting this latent distribution conditioned solely on the current visual observation and language instruction. The continuous latent plan is a sampled vector from these distributions that represents the high-level strategy for the task. We apply a KL-divergence between the proposal and recognition distributions for mapping from the current observation to future actions. In detail, we set the hidden size of the plan proposal and recognition modules to 2048, and the balance coefficient of the KL loss is set to 0.8.

\subsection{HULC}

HULC \cite{mees2022matters} conducts an extensive study of the critical challenges in the robot manipulation task and makes several improvements compared with the above MCIL. First, it learns discrete representations rather than a continuous vector for a natural fit for complex reasoning, planning, and predictive learning. Then, it improves the plan proposal and plan recognition networks by leveraging the Transformer \cite{vaswani2017attention} architecture to model the long-range and hierarchical robotic controls. In detail, we initialize a 32$\times$32 discrete space to simulate hierarchical intentions during human activity. Then, we adopt the Transformer-based plan proposal and plan recognition networks to model long-range dependencies, and the corresponding hidden size is set to 2048. Finally, we utilize an MLP layer to tokenize the learned discrete plan representations for the subsequent reasoning and forecasting. As in the MCIL approach, the KL-divergence between the proposal and recognition distributions is also performed, and the balancing coefficient is set to 0.8.

\subsection{GR-1}

GR-1 \cite{wuunleashing} is a representative VRM (also known as Vision-Language-Action (VLA)) work that leverages human data to facilitate downstream robot manipulation. The framework of GR-1 is a GPT-style model, which is first pre-trained on human Ego videos in Ego4D \cite{grauman2022ego4d} with a frame-wise generation objective. Then, it is finetuned on the downstream robotic data and achieves impressive performance on robot manipulation tasks. In this paper, GR-1 takes multimodal cues (i.e. Ego visual observations, language instruction, and hand pose states) as input with two major training objectives. On the one hand, the framework is optimized by standard smoothL1 and BCE losses to predict 3D hand joint positions with validity masks. On the other hand, we feed the decoded queries into a multi-block MAE \cite{he2022masked} decoder to reconstruct the future third Ego frame relative to the current timestep by an MSE loss following the officially released code. The depth of the MAE decoder is set to 2, and the balancing coefficient of the reconstruction loss is set to 1.0.

\subsection{AR-VRM}

AR-VRM \cite{yang2025ar} is a recent method based on a pretrain-finetune paradigm. To address the problem of implicitly transferring knowledge from humans to robots, it is first pretrained by predicting the human hand keypoints in the Ego view. Then, it is finetuned on robotic data with retrieval and analogical reasoning strategies for human-to-robot transfer and achieves remarkable performance. In this paper, we re-implement this approach and further improve it by incorporating the paired exocentric (Exo) demonstration videos as in our Exo2EgoPose for cross-view reasoning. In detail, following the officially released code, we implement the analogical reasoning by adopting a multi-head Transformer with cross-attention blocks. Then, we treat the learned Ego pose features as queries and the additionally extracted Exo features as memory, which are fed into the multi-head analogical reasoning Transformer. Finally, we set a learnable balancing parameter to control the weight of the analogical reasoning predictions. Specifically, the depth of the analogical reasoning Transformer is set to 2, and the hidden size is set to 384. The learnable balancing parameter is initialized to 1.0 following the officially released code.

\section{Details of Dataset Pre-processing}

We have introduced the detailed construction pipeline of our \textit{EgoMe-pose} benchmark in Section 4.1.1 of the main text. In this section, we present the detailed dataset pre-processing pipeline for the other two benchmarks (i.e., \textit{AssemblyHands} \cite{ohkawa2023assemblyhands} and \textit{Ego-Exo4D} \cite{grauman2024ego}).

\subsection{\textit{AssemblyHands} Benchmark}

\begin{figure}[!t]
\centering
\includegraphics[width=0.72\linewidth]{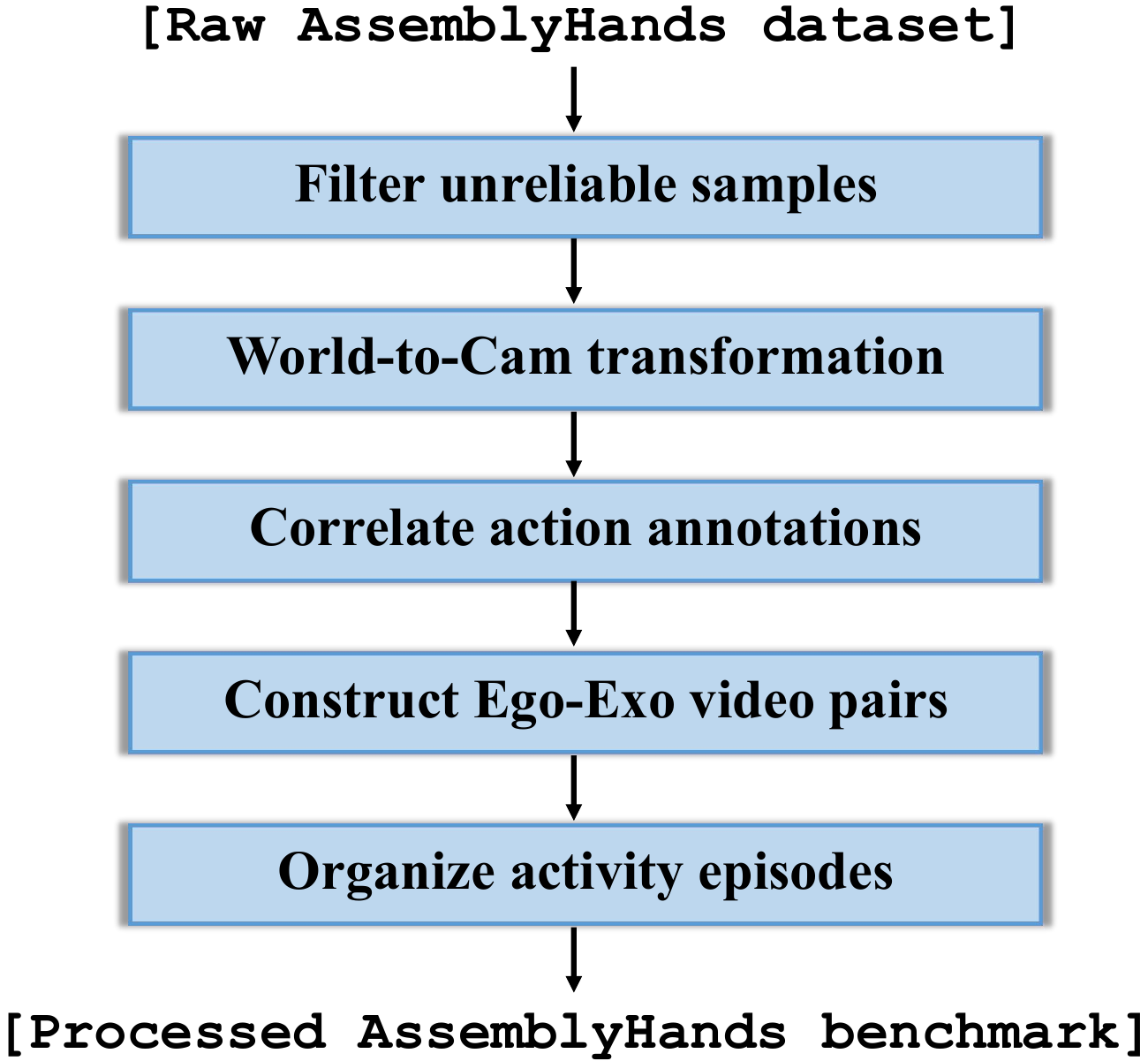}
\caption{The dataset pre-processing pipeline for the \textit{AssemblyHands} benchmark.}
\label{fig:a1}
\end{figure}

The AssemblyHands \cite{ohkawa2023assemblyhands} dataset provides precise 3D hand pose annotations based on the Assembly101 \cite{sener2022assembly101} dataset, which is designed for toy assembly scenarios. To adapt this dataset to our VL-EHPF task, we conduct thorough data pre-processing to construct a multimodal \textit{AssemblyHands} benchmark as shown in Fig. \ref{fig:a1}. The detailed data pre-processing pipeline is illustrated as follows:

\textbf{(1) Filter unreliable samples}: We observed that some annotation cases in the original AssemblyHands dataset suffered from missing hand joints. Therefore, we filter unreliable samples by retaining a frame's annotation if each hand contains more than 10 valid joints with a valid root joint. Otherwise, the hand pose annotation of the current frame is discarded.

\textbf{(2) World-to-Cam transformation}: The initial 3D hand pose annotations in AssemblyHands are labeled in world coordinates, while our task forecasts hand poses in the Ego camera coordinate system. Thus, we perform a world-to-camera coordinate transformation. Specifically, we use the official extrinsic parameters, including the rotation matrix $R$ and translation vector $T$ to convert each 3D joint from the world coordinate to the Ego camera coordinate.

\textbf{(3) Correlate action annotations}: Since the original AssemblyHands dataset lacks language annotations, we utilize the textual fine-grained action annotations in the Assembly101 dataset to serve as language instructions. Specifically, we correlate the Ego video frames, 3D hand pose annotations, and language instructions based on the unified frame IDs shared between the two datasets.

\textbf{(4) Construct Ego-Exo video pairs}: We further construct Ego-Exo video pairs for training our Exo2EgoPose framework. Thanks to the AssemblyHands dataset, which provides Ego videos with synchronized multiple-view Exo demonstrations, we randomly select one of the Exo views to form the Ego-Exo video pairs.

\textbf{(5) Organize activity episodes}: Finally, we organize multimodal data into activity episodes. Each episode contains all the Ego video frames, corresponding 3D hand poses, and language instructions for an entire activity. Since the original dataset does not provide a publicly available test set, we randomly sample 50$\%$ episodes from the \textit{val} set to construct the \textit{test} set.

\subsection{\textit{Ego-Exo4D} Benchmark}

\begin{figure}[!t]
\centering
\includegraphics[width=0.72\linewidth]{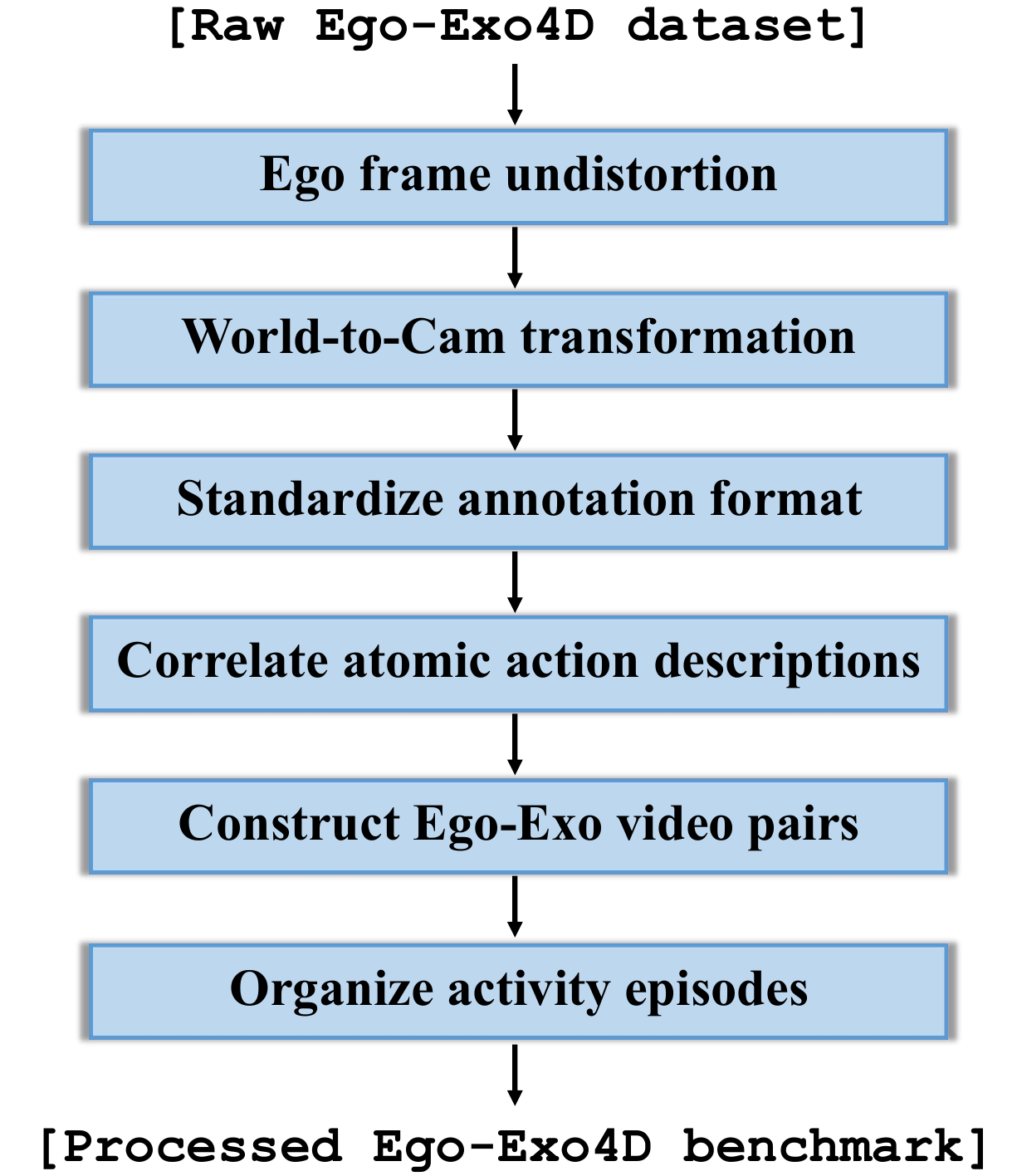}
\caption{The dataset pre-processing pipeline for the \textit{Ego-Exo4D} benchmark.}
\label{fig:a2}
\end{figure}

\begin{figure*}[!t]
\centering
\includegraphics[width=1.0\linewidth]{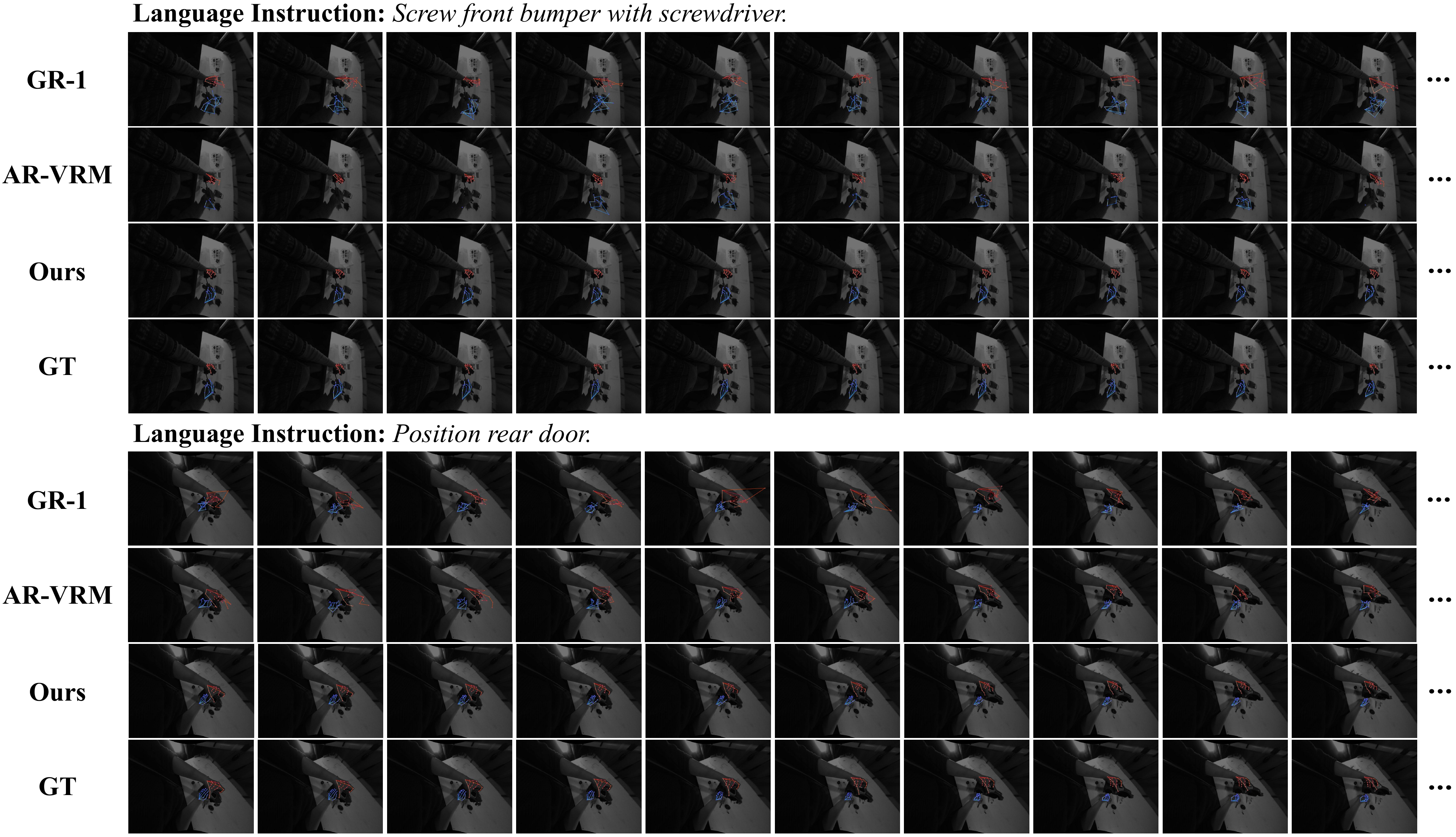}
\caption{Visualization results of the continuously forecasted Ego 3D hand poses for the next 10 frames of our Exo2EgoPose and comparison methods on the \textit{AssemblyHands} benchmark. (Due to the small scale of hands in the original Ego frames, \textcolor{red}{zooming in is recommended} for better visibility of pose details).}
\label{fig:a3}
\end{figure*}

\begin{figure*}[!t]
\centering
\includegraphics[width=1.0\linewidth]{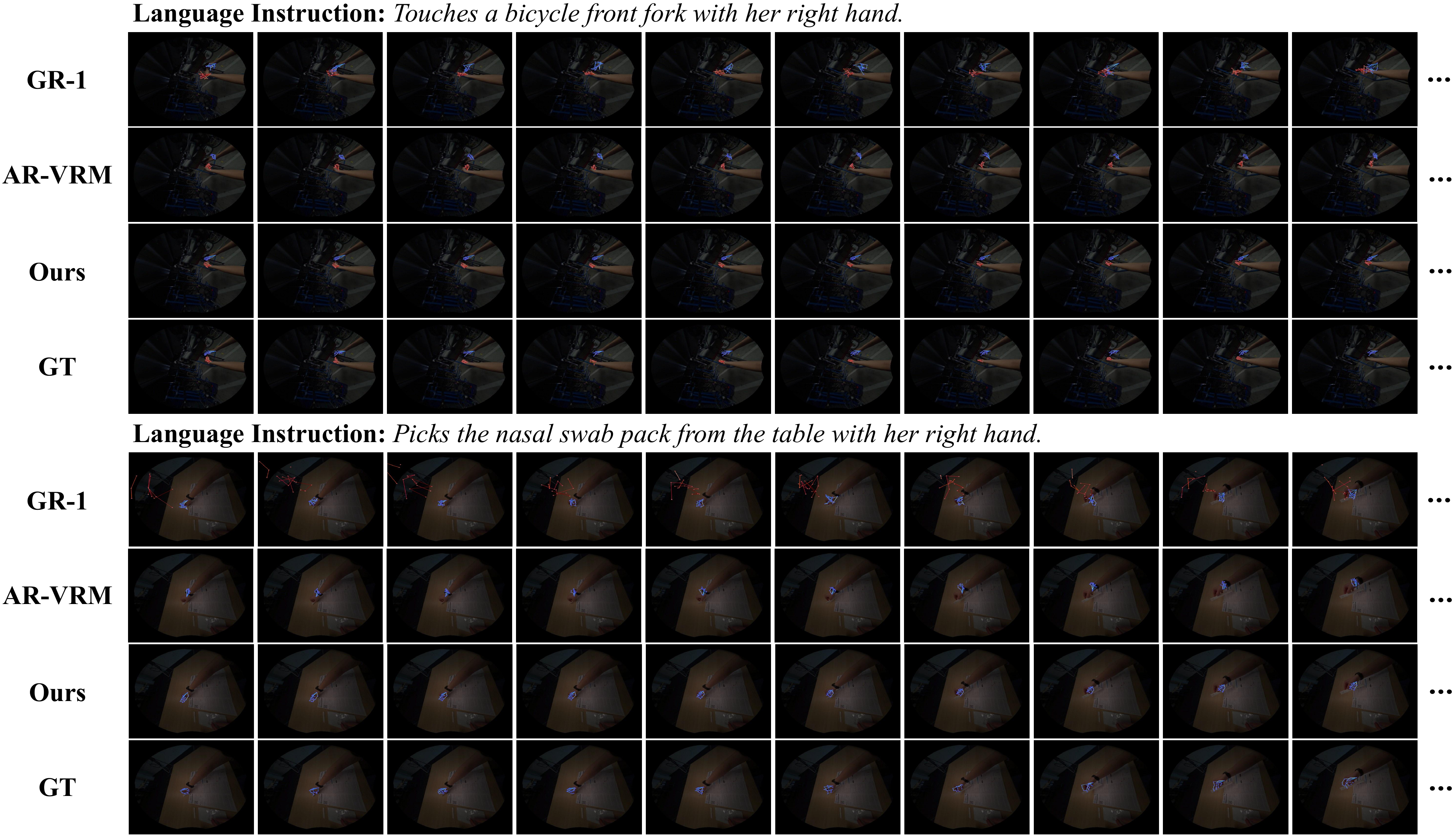}
\caption{Visualization results of the continuously forecasted Ego 3D hand poses for the next 10 frames of our Exo2EgoPose and comparison methods on the \textit{Ego-Exo4D} benchmark. (Due to the tiny scale of hands in the original Ego frames, \textcolor{red}{zooming in is recommended} for better visibility of pose details).}
\label{fig:a4}
\end{figure*}

\begin{figure*}[!t]
\centering
\includegraphics[width=1.0\linewidth]{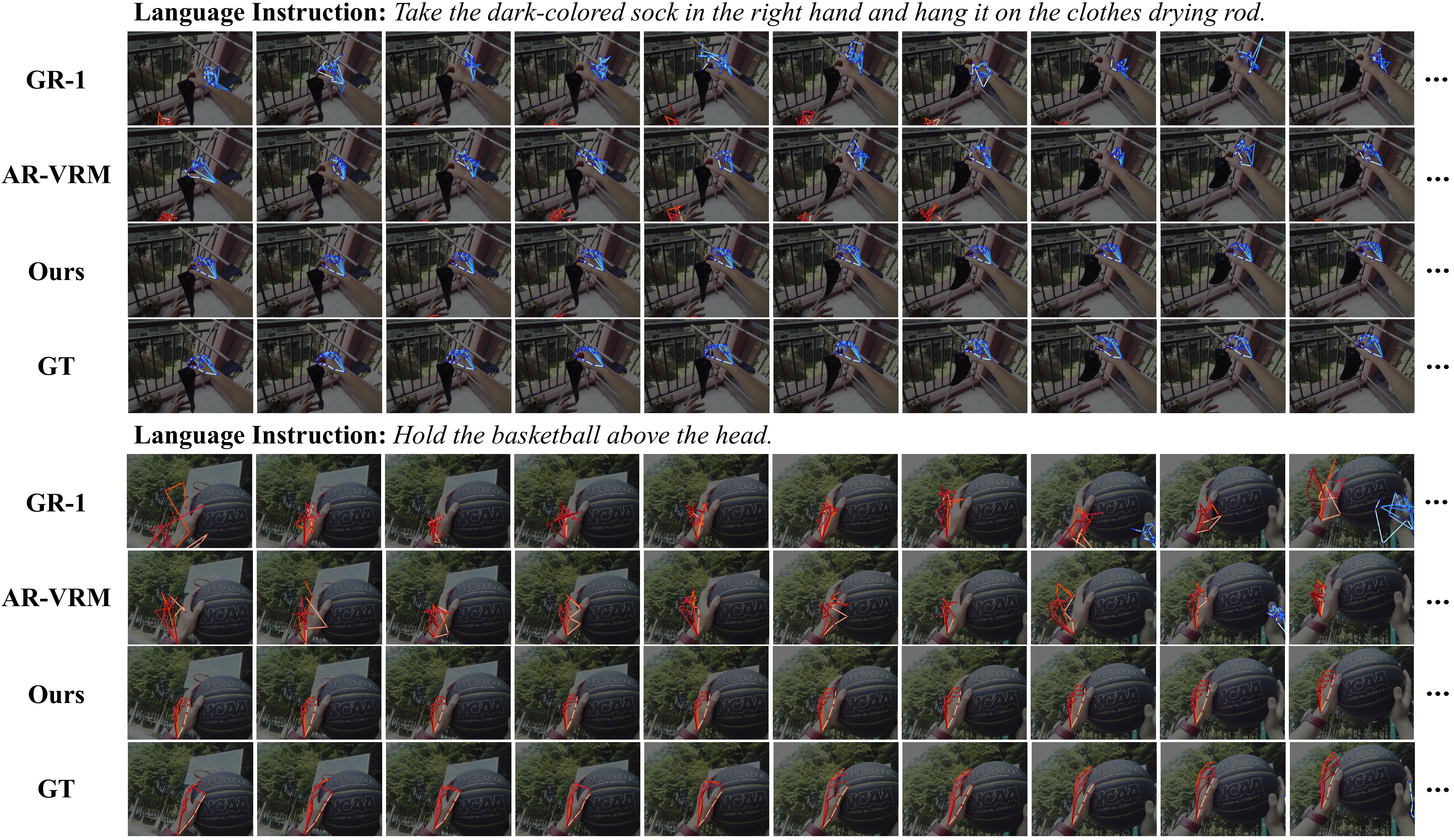}
\caption{Visualization results of the continuously forecasted Ego 3D hand poses for the next 10 frames of our Exo2EgoPose and comparison method on the \textit{EgoMe-pose} benchmark.}
\label{fig:a5}
\end{figure*}

The Ego-Exo4D \cite{grauman2024ego} dataset is the largest Ego-Exo dataset with diverse annotations such as 3D hand poses, atomic action descriptions, expert commentary, and so on. To construct the \textit{Ego-Exo4D} benchmark suitable for our VL-EHPF task, we perform data pre-processing as shown in Fig. \ref{fig:a2}. The detailed pipeline is as follows:

\textbf{(1) Ego frame undistortion}: The Ego videos in the Ego-Exo4D dataset are captured by open-source Aria glasses with inherent image distortion. Thanks to the fact that the dataset provides \textit{.vrs} files containing Ego camera parameters, we perform image undistortion on all involved Ego video frames according to the official camera parameters and the undistortion toolbox.

\textbf{(2) World-to-Cam transformation}: Similar to the \textit{AssemblyHands} benchmark, we perform the world-to-camera transformation for the Ego 3D hand pose annotations in the Ego-Exo4D dataset. We leverage the official camera extrinsic parameters for the above coordinate system transformation for each hand joint.

\textbf{(3) Standardize annotation format}: Since the hand joint indexing rule of the pose annotations in the Ego-Exo4D dataset is inconsistent with the universal joint indexing rule \cite{ohkawa2023assemblyhands,moon2020interhand2}, we perform annotation standardization by mapping all hand pose annotations to ensure consistency with the widely-used hand joint protocol.

\textbf{(4) Correlate atomic action descriptions}: To assign language instructions to the video clips with poses, we exploit the textual annotations in the dataset. We correlate the Ego clips and poses with the atomic action description annotations in Ego-Exo4D within a pre-defined temporal window. In detail, we define the window as the interval from 0.5 seconds before to 1.0 seconds after the atomic action timestamp, with the requirement that every frame within this window has valid pose annotations.

\textbf{(5) Construct Ego-Exo video pairs}: Similar to the \textit{AssemblyHands} benchmark, we randomly select one of the synchronized Exo views as paired Exo demonstration videos to construct the Ego-Exo video pairs to train our Exo2EgoPose framework.

\textbf{(6) Organize activity episodes}: Finally, we organize the above data into ``video-language-pose" triplets. Due to the absence of the publicly released \textit{test} set of the Ego-Exo4D hand pose benchmark, we randomly sample 50$\%$ of the episodes of the \textit{val} split to construct the \textit{test} set of our processed \textit{Ego-Exo4D} benchmark.

\section{More Visualization Results}

In this section, we show more visualization results of the continuously forecasted Ego 3D hand poses for the next 10 frames of our Exo2EgoPose and two recent comparison methods (i.e., GR-1 \cite{wuunleashing} and AR-VRM \cite{yang2025ar}) on the existing \textit{AssemblyHands} \cite{ohkawa2023assemblyhands}, \textit{Ego-Exo4D} \cite{grauman2024ego} benchmarks, and our newly constructed \textit{EgoMe-pose} benchmark built upon the EgoMe \cite{qiu2025egome} dataset. These visualizations comprehensively and intuitively demonstrate the effectiveness of the proposed Exo2EgoPose approach.

\subsection{Qualitative Results on \textit{AssemblyHands}}

In Fig. \ref{fig:a3}, we show two visualization examples on the \textit{AssemblyHands} \cite{ohkawa2023assemblyhands} benchmark of the Ego 3D hand poses for the next 10 consecutive frames, which are forecasted by our Exo2EgoPose and other comparison methods \cite{wuunleashing,yang2025ar}. \textit{AssemblyHands} is a widely used benchmark that mainly contains activities of assembling toy cars. In this benchmark, hands typically occupy only small regions within the frame, which requires fine-grained hand-object interaction understanding and modeling. In the upper example, the language instruction is ``Screw front bumper with screwdriver", which requires the left hand to hold the toy car and the right hand to operate the screwdriver. In the first row, GR-1 outputs and distinguishes left and right hands. However, the hand structures are cluttered and cannot present the screwing process. In the second row, AR-VRM forecasts coarse hand structures, while the joints are inconsistent with those of the ground-truths. Moreover, the poses of the right hand are missing in several frames. In contrast, in the third row, our Exo2EgoPose method generates precise hand poses, where the left hand holds the object and the right hand rotates the screwdriver slowly. It demonstrates that our method accurately understands this activity and forecasts the fine-level hand actions.

In the bottom example, the subject aims to ``Position rear door", where the left hand lifts the toy car and the right hand adjusts its gesture to prepare to position the rear door. In the first row, GR-1 fails to generate plausible structures, as some skeletons are unexpectedly elongated. In the second row, AR-VRM roughly predicts coarse but seemingly plausible hand structures that still largely deviate from the ground-truths. Specifically, the hand poses are distorted in many frames and do not adequately reflect the referred action. In the third row, the proposed Exo2EgoPose method outputs precise hand poses with plausible hand structures and accurately reflects the current fine-grained atomic action.

\subsection{Qualitative Results on \textit{Ego-Exo4D}}

In Fig. \ref{fig:a4}, we visualize two examples of the next 10 consecutive Ego 3D hand poses on the \textit{Ego-Exo4D} \cite{grauman2024ego} benchmark, which are forecasted by GR-1 \cite{wuunleashing}, AR-VRM \cite{yang2025ar}, and our Exo2EgoPose. \textit{Ego-Exo4D} is currently the largest multi-view Ego-Exo dataset with detailed annotations and covers diverse real-world activities. Since the videos in \textit{Ego-Exo4D} are captured by fish-eye Project Aria glasses \cite{engel2023project}, the hand areas become very small after image undistortion. Please zoom in on Fig. \ref{fig:a4} for better visualization of hand poses. In the upper example, the subject is going to ``Touch a bicycle front fork with her right hand" in the next several timestamps. GR-1 in the first row predicts messy hand joint positions. In the second row, AR-VRM can form complete structures for both hands. However, the overall orientations of the predicted hands are significantly different from those of the ground-truths. Finally, despite a few valid joints being missing in the ground-truth pose annotations, our method still generates precise Ego hand poses for the process of touching the bicycle.

In the bottom case, the given language instruction is ``Picks the nasal swab pack from the table with her right hand". In the first row, GR-1 incorrectly generates false-positive cluttered joints of the left hand, which is not involved in the entire activity process. In the second row, AR-VRM fails to comprehend the ``picking the nasal swab pack" action and generates hand poses remarkably different from the ground-truth poses. On the contrary, in the third row, our Exo2EgoPose accurately understands this process and outputs the correct hand poses in the Ego view, where the right hand gradually contracts for grasping the object of interest. The above visualization results further demonstrate the effectiveness of our ideology that reconstructs the Exo demonstrations at different levels and then leverages the Exo guidance to perform global-to-local feature modulation for accurate Ego 3D hand pose forecasting.

\subsection{Qualitative Results on \textit{EgoMe-pose}}

In Fig. \ref{fig:a5}, we show two visualization examples of continuously forecasted Ego 3D hand poses for the future 10 frames by the proposed Exo2EgoPose and other comparison methods (i.e., GR-1 \cite{wuunleashing} and AR-VRM \cite{yang2025ar}) on our newly constructed \textit{EgoMe-pose} benchmark. \textit{EgoMe-pose} is constructed based on the recent EgoMe \cite{qiu2025egome} dataset, which contains paired yet asynchronous Ego and Exo videos captured by observers and followers with dynamic camera motions in diverse scenarios. The visualization strategy is consistent with that applied in the main text. In the upper example, the hand poses of ``Take the dark-colored sock in the right hand and hang it on the clothes drying rod" are required to be forecasted. In the first row, GR-1 predicts cluttered hand poses and false-positive joints of the irrelevant left hand. It demonstrates that GR-1 fails to understand the process of ``picking up a sock with the right hand" during the holistic hanging activity. In the second row, the hand poses forecasted by AR-VRM roughly form the hand structure and capture the posture of picking up a sock. Nevertheless, the detailed positions of each hand joint deviate significantly from the ground-truths. In contrast, our Exo2EgoPose in the third row generates plausible hand structures and gestures with precise hand joint positions and reasonable dynamics.

The bottom example focuses on the scenario of playing basketball, the subject is going to ``Hold the basketball above the head" and prepares to shoot. Due to the limited field-of-view and highly dynamic motions during playing basketball in the Ego view, GR-1 and AR-VRM in the first and second rows both struggle to understand the hand states during shooting a basketball (i.e., the hand joints need to wrap around the basketball). Thus, they both predict incorrect hand joint positions, which are inconsistent with the ground-truth hand poses. Conversely, in the third row, our Exo2EgoPose incorporates holistic and stable Exo demonstration videos as guidance for progressive global-to-local feature modulation, resulting in accurate forecasted Ego 3D hand poses for the referred future action.











\end{document}